\PassOptionsToPackage{usenames, dvipsnames}{xcolor}
\documentclass[a4paper, 11pt]{preprint}
\pdfoutput=1

\usepackage[usenames, dvipsnames]{xcolor}

\usepackage{tikz}

\usepackage{enumitem}
\usepackage{amsfonts}
\usepackage{amsmath}
\usepackage{hyperref}

\usepackage{natbib}
\usepackage{caption}
\usepackage[font=small]{subcaption}
\usepackage{url}

\usepackage{mathtools}
\usepackage{hyperref}
\usepackage{booktabs}
\usepackage{siunitx}

\usepackage{graphicx}

\usepackage{newtxtext}
\usepackage[subscriptcorrection]{newtxmath}

\usepackage[plain,noend]{algorithm2e}

\makeatletter
\renewcommand{\algocf@captiontext}[2]{#1\algocf@typo. \AlCapFnt{}#2}

\def\@algocf@capt@plain{top}
\renewcommand{\algocf@makecaption}[2]{%
  \addtolength{\hsize}{\algomargin}%
  \sbox\@tempboxa{\algocf@captiontext{#1}{#2}}%
  \ifdim\wd\@tempboxa >\hsize%     % if caption is longer than a line
    \hskip .5\algomargin%
    \parbox[t]{\hsize}{\algocf@captiontext{#1}{#2}}% then caption is not centered
  \else%
    \global\@minipagefalse%
    \hbox to\hsize{\box\@tempboxa}% else caption is centered
  \fi%
  \addtolength{\hsize}{-\algomargin}%
}
\makeatother

\newcommand\independent{\protect\mathpalette{\protect\independenT}{\perp}}
\def\independenT#1#2{\mathrel{\rlap{$#1#2$}\mkern2mu{#1#2}}}

\DeclareMathOperator{\G}{\mathcal{G}}
\DeclareMathOperator{\pa}{pa}
\DeclareMathOperator{\cpa}{pa}

\DeclareMathOperator{\ms}{ms}
\newcommand{\vX}{{\bf X}}
\newcommand{\vx}{{\bf x}}
\newcommand{\vy}{{\bf y}}

\DeclarePairedDelimiterX{\klx}[2]{(}{)}{%
	#1\;\delimsize\|\;#2%
}
\newcommand{\kl}{D_\mathrm{KL}\klx}

\title{Scalable Structure Learning for Sparse Context-Specific Systems}
\author{Felix Leopoldo Rios, Alex Markham and Liam Solus}

\address[F.~L.~Rios]{Department of Mathematics, KTH Royal Institute of Technology, Sweden}
\email{flrios@kth.se}

\address[A.~Markham]{Department of Mathematics, KTH Royal Institute of Technology, Sweden}
\email{markham@kth.se}

\address[L.~Solus]{Department of Mathematics, KTH Royal Institute of Technology, Sweden}
\email{solus@kth.se}
\date{\today}

\keywords{%
  structure learning,
  causal discovery,
  graphical model,
  context-specific graphical model,
  directed acyclic graph,
  CStree,
  labeled directed acyclic graph,
  staged tree.}

\begin{document}

\begin{abstract}
	Several approaches to graphically representing context-specific relations among jointly distributed categorical variables have been proposed, along with structure learning algorithms.
	While existing optimization-based methods have limited scalability due to the large number of context-specific models, the constraint-based methods are more prone to error than even constraint-based directed acyclic graph learning algorithms since more relations must be tested.
	We present an algorithm for learning context-specific models that scales to hundreds of variables.
	Scalable learning is achieved through a combination of an order-based Markov chain Monte-Carlo search and a novel, context-specific sparsity assumption that is analogous to those typically invoked for directed acyclic graphical models.
	Unlike previous Markov chain Monte-Carlo search methods, our Markov chain is guaranteed to have the true posterior of the variable orderings as the stationary distribution.
	To implement the method, we solve a first case of an open problem recently posed by Alon and Balogh.
	Future work solving increasingly general instances of this problem would allow our methods to learn increasingly dense models.
	The method is shown to perform well on synthetic data and real world examples, in terms of both accuracy and scalability.
\end{abstract}

\maketitle

\section{Introduction}

This paper examines the problem of structure learning (sometimes referred to as causal discovery) for categorical data where one wishes to infer a compact, graphical representation of context-specific conditional independence relations in the data-generating distribution.
Given a joint distribution $\vX = (X_1,\ldots, X_p)$ and disjoint subsets $A,B,C\subseteq[p]:=\{1,\ldots, p\}$ with $A$ and $B$ nonempty, $\vX_A$ is said to be conditionally independent of $\vX_B$ given $\vX_C$, written $\vX_A \independent \vX_B \mid \vX_C$, if $ P(\vx_A \mid \vx_B, \vx_C) = P(\vx_A \mid \vx_C) $ for all marginal outcomes $\vx_A, \vx_B$ and $\vx_C$.
Estimating CI relations entailed by jointly distributed categorical variables is well-known to reduce the number of parameters needed when performing inference tasks \citep{koller2009probabilistic}.
This observation is one main motivation for estimating a directed acyclic graph representation of a distribution.
Given a directed acyclic graph $\G = ([p], E)$ with node set $[p]$ and edge set $E$, the distribution $\vX$ is \emph{Markov} to $\G$ if
\begin{equation}
	\label{eqn:factorization}
	P(\vx) = \prod_{i=1}^p P(x_i \mid \vx_{\pa_{\G}(i)}) \,\,\,\, \mbox{for all outcomes $\vx = (x_1,\ldots,x_p)$,}
\end{equation}
where $\pa_{\G}(i) = \{k\in [p] : k\rightarrow i\in E\}$ denotes the set of \emph{parents} of $i$ in $\G$.
The factorization reveals how conditional independence relations entailed by the distribution reduce the number of parameters: When $\G$ is sparse, the parent sets are small, reducing the number of parameters needed to represent the distribution as a product of conditional probabilities.

A directed acyclic graph representation (i.e., a directed acyclic graph $\G$ to which the distribution is Markov) is thus a compact representation of the conditional probability table.
Instead of storing the conditional probabilities $P(x_i \mid \vx_{[i-1]})$ for all $i\in[p]$ and all outcomes $\vx_{[i-1]}$, we store only enough to represent the factors $P(x_i \mid \vx_{\pa_{\G}(i)})$ for all $i$.
When $\G$ is sparse, this significantly reduces storage requirements for representing the distribution, and allows for speedy inference computations via message-passing \citep{lauritzen1988local, shafer1990probability}.
However, a directed acyclic graph representation may obscure important relations that only hold for certain outcomes of a conditioning set.

For disjoint subsets $A,B,C,S\subseteq [p]$ we say $\vX_A$ is \emph{conditionally independent of} $\vX_B$ \emph{given} $\vX_C$ \emph{in the context} $\vX_S = \vx_S$ if
\begin{equation}
	\label{eqn:CSIdefinition}
	P(\vx_A | \vx_B, \vx_C, \vx_S) = P(\vx_A | \vx_C, \vx_S) \, \, \, \mbox{for all} \, \, \vx_A, \vx_B, \vx_C.
\end{equation}
We denote the relation~\eqref{eqn:CSIdefinition} by $\vX_A \independent \vX_B | \vX_C, \vX_S = \vx_S$ and call it a \emph{context-specific conditional independence relation} or \emph{CSI relation}.
Context-specific conditional independence relations arise naturally in a variety of modeling and decision-making problems \citep{poole2003exploiting}, such as the following.

\begin{example}
	(Adapted from \citep[Example 5]{poole2003exploiting}.)
	\label{ex:CSI}
	When a child checks into the emergency room, the hospital staff may want to determine if they are a carrier of chicken pox, so steps can be taken to avoid spreading the disease.
	If the child has not recently been exposed to chicken pox, they are unlikely to be a carrier.
	Hence, we may assume carrier status is independent of all other background information, given that they have not been exposed.
	On the other hand, given that they have been exposed and have no previous diagnosis then they are likely a carrier regardless of other background information.
	Similarly, given that the child has a previous diagnosis, recent exposure may be independent of other background information, as less precautionary steps may be taken to avoid another exposure.
	Letting $X_4,X_3,X_2,X_1$ denote, respectively, carrier, recent exposure, previous diagnosis, and some background covariate assumed to be independent of $X_2$, we obtain the following context-specific conditional independence model \[ \{ X_4\independent \vX_{1,2} \mid X_3 = \mbox{no},\, X_4 \independent X_1 \mid \vX_{2,3} = (\mbox{no}, \mbox{yes}),\, X_3 \independent X_1 \mid X_2 = \mbox{yes},\, X_2 \independent X_1 \}.
	\]
\end{example}
Example~\ref{ex:CSI} highlights how knowledge of context-specific conditional independence relations may play an important role in predictive models for decision-making.
Since these potentially important relations, which are overlooked by directed acyclic graph representations of data, may exist in much larger systems, there is a general need for compact representations of context-specific conditional independence relations that can be learned efficiently and at scale.

Encoding context-specific conditional independence relations in the conditional probability table captures more information than the conditional independence relations of directed acyclic graphical models at the price of additional computation and storage requirements, making efficient and scalable learning a challenging task.
On the other hand, the additional context-specific information can also boost efficiency in inference \citep{boutilier2013context}.
By imposing sparsity requirements that limit the size of the sets of variables $S$ defining the relevant contexts, one can strike a balance between additional detail and computational efficiency.
In this paper, we present an algorithm for learning a graphical representation of sparse context-specific models that achieves this balance.

While a variety of optimization-based structure learning algorithms for context-specific models exist \citep{hyttinen2018structure, leonelli2023context, pensar2015labeled}, many of these methods strive for a broad generality in the context-specific relations they can capture at the expense of only being able to learn models on few variables ($p \ll 100$).
Constraint-based methods that generalize the PC algorithm for directed acyclic graphs by analogous testing of context-specific relations were also studied \citep{hyttinen2018structure}.
While faster, these algorithms risk even more error than constraint-based directed acyclic graph algorithms due to the increased number of constraints.

We present a structure learning algorithm that combines a Markov chain Monte-Carlo sampling, some exact optimization, and, optionally, some non-context-specific conditional independence testing.
This optional constraint-based phase tests no more conditional independence relations than a constraint-based directed acyclic graph learner, limiting potential error propogation.

Markov chain Monte-Carlo methods are highly effective in the directed acyclic graph learning regime according to the recent \texttt{Benchpress} platform for comparing structure learning algorithms \citep{rios2021benchpress}.
Unlike the existing methods, our Markov chain Monte-Carlo sampler estimates the full posterior distribution of the variable orderings.
To implement this sampler, we solve a special case of a combinatorial problem posed by \citet{alon2023partitioning}.
To the best of our knowledge, our solution is the first settled case of this problem, which is well-known to be difficult in its fullest generality.
We describe how future work that incrementally generalizes our result (Theorem~\ref{thm:enumeratingstagings}) on this problem allows for our method to learn increasingly dense models.

Finally, we introduce a novel context-specific sparsity assumption that allows us to efficiently perform exact optimization over all models with the optimal variable ordering.
The algorithm scales ($p \gg 100$) while returning a sparse graphical representation of the conditional probability table with the same advantages for storage and inference as sparse directed acyclic graphical models.
Scalability is demonstrated through both complexity bounds (Theorem~\ref{thm:loscorecomp}) and evaluation on synthetic data and real world examples.
The experiments also show the method achieving a high level of accuracy.
Our implementation and experimental analysis are publicly available as part of an open-source Python package for context-specific statistical learning methods: \href{https://cstrees.readthedocs.io/en/latest/index.html}{CSlearn}.

\section{Connections to Related Work}\label{sec:relatedwork}

Several models for context-specific relations exist, ranging from relatively minimal assumptions yielding subtle extensions of directed acyclic graphical models to broad attempts at capturing many context-specific relations.
The former models include \emph{Bayesian multinets} \citep{geiger1996knowledge} and \emph{similarity networks} \citep{heckerman1990probabilistic} which introduce a single variable to distinguish between contexts, called the \emph{hypothesis variable} or \emph{distinguished node}.
These models allow for context-specific conditional independence relations where the contexts are limited to the scope of the hypothesis variable.
They possess many of the efficiencies of directed acyclic graphical models but are limited in the context-specific relations they capture.

At the other extreme are \emph{staged trees}, introduced by \cite{smith2008conditional}, capable of encoding a wealth of context-specific relations.
However, the graphical representation of a staged tree typically lacks interpretability since the size of the graph grows on the order of $2^p$ for even binary variables.
\emph{Chain Event Graphs} \citep{smith2008conditional} reduce the staged tree representation, but can still be difficult to interpret.

To overcome scalability problems, subfamilies of staged trees were studied and associated structure learning methods were proposed.
\citet{leonelli2022highly} proposed a method that first estimates a directed acyclic graph with bounded in-degree using classical causal discovery methods and then greedily optimizes the \emph{staging}, which represents the context-specific relations in the model.
The method is relatively efficient due to the sparsity constraints imposed by bounds on the parent sets in the graph.
On the other hand, the learned staged tree may still be a very large and complex graph, making it difficult to interpret.
The models learned by our method admit a more compact representation, akin to a directed acyclic graph.

Other staged tree algorithms instead limit variable ordering without the use of a directed acyclic graph \citep{strong2022scalable}.
However, these methods inevitably use a greedy search to learn context-specific relations which risks error that is avoided by our exact optimization approach to this step.
Some staged tree algorithms \citep{leonelli2023context} achieve efficiency by only searching over a fixed variable ordering, which is subject to misspecification.
Performing a Markov Chain Monte-Carlo search on the order space is known to be effective in DAG learning \citep{friedman2003being, kuipers2022efficient}.
In our context, it allows us to efficiently estimate an optimal variable ordering from posterior samples.
This Bayesian approach also allows for quantifying the uncertainty in our estimate.

Between the two extremes are the \emph{Labeled directed acyclic graphs} (LDAGs) \citep{pensar2015labeled} that are more interpretable than staged trees but also capture more context-specific information than similarity networks.
LDAGs are context-specific models that encode pairwise context-specific conditional independence relations $X_i \independent X_j | \vX_{\pa_{\G}(i)\setminus j} = \vx_{\pa_{\G}(i)\setminus j}$ where $j\rightarrow i$ is an edge in a directed acyclic graph representation of the distribution; i.e., they encode context-specific relations that represent a single edge vanishing for a specified context of the other parents of $i$.
Such models have proven useful, as they are more amenable to both structure learning \citep{hyttinen2018structure} and causal inference.
In the latter case, LDAGs help capture additional causal relations from only observational data \citep{tikka2019identifying} by identifying additional \emph{essential arrows} \citep{andersson1997characterization} in the directed acyclic graph.

\citet{hyttinen2018structure} explored constraint-based and exact optimization-based methods for learning LDAGs.
While the constraint-based methods can scale under sparsity constraints, the algorithms run analogous to the PC algorithm \citep{spirtes1991algorithm} but test CSI relations over all possible contexts.
This requires more tests that may lead to errors in addition to those of the classical PC algorithm.
Our proposed method need not test anymore CI relations than classical PC, limiting the possibility for error.
\citet{hyttinen2018structure} also show how exact optimization methods can be done using local score computations for LDAGs, but they note that these methods are generally infeasbile for distributions with more than four variables (although they include an example on $37$ nodes with variables on up to four states with a rather impressive run time of only seven hours on a modern desktop computer).
\citet{pensar2015labeled} present a non-reversible Markov chain approach for learning LDAGs that blends stochastic and greedy search.
This method potentially achieves higher accuracy over the constraint-based methods.
However, \cite{hyttinen2018structure} note that the blending of stochastic and greedy search necessitates careful parameter tuning to avoid overfitting and to obtain accuracy gains over constraint-based methods.

A potential downside of LDAGs is that the pairwise context-specifc relations defining the model always have contexts $\vx_{S}$ where $|S| = |\pa_{\mathcal{G}}(i)| -1$, meaning that sparsity of the model is only easily controlled at the level of graph sparsity.
This perspective may overlook opportunities for flexibility in structure learning that utilize sparsity bounds based directly on the context-specific nature of the model; e.g., as a function of $|S|$ for general contexts $\vx_S$, rather than only those arising as outcomes of parents in a graph.
Our proposed methods allow for the combination of such novel sparsity sparsity constraints (Assumption~\ref{ass:CS}) together with classical graph sparsity (Assumption~\ref{ass:DAG}) already used in LDAG learning \citep{hyttinen2018structure, pensar2015labeled}.

Recently, a family of submodels of LDAGs, called \emph{CStrees}, were isolated and studied by \citet{duarte2021representation}.
Instead of pairwise context-specific conditional independence relations, CStrees are defined via a relaxation of the factorization in \eqref{eqn:factorization} (see Section~\ref{sec:cstrees}).
CStrees are submodels of LDAGs and all directed acyclic graphical models are CStree models.
Hence, CStrees possess the desirable features of LDAGs while being somewhat closer to directed acyclic graphical models.
Since CStrees are defined according to a factorization, they admit natural opportunities to impose the aforementioned novel context-specific sparsity constraints.
These sparsity constraints allow for model enumeration (Theorem~\ref{thm:enumeratingstagings}), which is necessary for computing model scores typically used in Monte-Carlo search algorithms.
Our enumeration appears to be the first solved, nontrivial case of a combinatorial problem posed by \citet{alon2023partitioning}, and, as we will see, solving increasingly general cases corresponds to relaxing our context-specific sparsity constraint.
In particular, future work on this problem will have immediate consequences for the inference of interpretable representations of context-specific structure in categorical data.

For these reasons, our proposed structure learning algorithm is designed to estimate sparse CStree models, while being easily adaptable to denser models when provided with generalizations of Theorem~\ref{thm:enumeratingstagings}.
Specifically, in Section~\ref{subsec:alg}, we take advantage of the features of CStrees to give a fast structure learning method whose estimates have the desirable properties of LDAGs while also capturing more information than a directed acyclic graph.

\section{Models}\label{sec:cstrees}
The models we study are a subfamily of LDAGs called CStrees.
They admit the CStree representation of \citet{duarte2021representation}, the staged tree representation of \citet{smith2008conditional} as well as the LDAG representation of \citet{pensar2015labeled}.
While LDAGs will be used to present an interpretable model representation, the CStree structure will be used to achieve fast structure learning.

A CStree model is defined for a collection of jointly distributed categorical variables $\vX = (X_1,\ldots, X_p)$, and it is simply an ordered pair $\mathcal{T} = (\pi, \mathbf{s})$ where $\pi$ is a variable ordering and $\mathbf{s}$ is a collection of sets.
Suppose $X_i$ has state space $[d_i]$ for $d_1,\ldots, d_p > 1$ and $\vX$ assumes the state space $\mathcal{X} = \prod_{i=1}^p[d_i]$.
For $S\subseteq [p]$ we let $\vX_S = (X_i : i\in S)$ denote the marginal distribution on the variables with indices in $S$, and we denote its state space with $\mathcal{X}_S$.
Let $\pi = \pi_1\cdots\pi_p$ denote a total ordering of the indices $[p]$ and consider CSI relations of the form
\begin{equation}
	\label{eqn:definingrelation}
	X_{\pi_i} \independent \vX_{[\pi_1:\pi_{i-1}]\setminus S} \mid \vX_S = \vx_S
\end{equation}
for some $\vx_S \in \mathcal{X}_S$ where $S\subseteq [\pi_1:\pi_{i-1}]:= \{\pi_1,\ldots, \pi_{i-1}\}$.
To this relation, we associate the set $\mathcal{S}_{\pi,i}(\vx_S)$ of all marginal outcomes $\vx_{\pi_1:\pi_{i-1}}\in\mathcal{X}_{[\pi_1:\pi_{i-1}]}$ that equal the context $\vx_S$ when restricted to the values with indices in $S$.
We allow for $\mathcal{S}_{\pi,i}(\vx_S)$ with $S = [\pi_1:\pi_{i-1}]$.
In this case, $\mathcal{S}_{\pi,i}(\vx_{\pi_1:\pi_{i-1}}) = \{\vx_{\pi_1:\pi_{i-1}}\}$ is a singleton corresponding to the relation $X_{\pi_i} \independent \emptyset \mid \vX_{\pi_1:\pi_{i-1}} = \vx_{\pi_1:\pi_{i-1}}$, meaning the conditional distribution $X_{\pi_i} \mid \vX_{\pi_1:\pi_{i-1}} = \vx_{\pi_1:\pi_{i-1}}$ may depend on the outcome $\vx_{\pi_1:\pi_{i-1}}$ of all preceding variables.

Let $\mathcal{C}_{\pi,i}$ denote a set of context-specific conditional independence relations of the form~\eqref{eqn:definingrelation} for a given $i\in[p]$ and the ordering $\pi$.
Consider $\mathcal{C}_{\mathcal{T}} = \mathcal{C}_{\pi,1}\cup \cdots \cup \mathcal{C}_{\pi,p}$, with one set $\mathcal{C}_{\pi,i}$ for each $i\in[p]$.
We obtain a set of distributions defined by the relations in $\mathcal{C}_{\mathcal{T}}$:
\begin{displaymath}
	\mathcal{M}(\mathcal{T}) = \{\vX : \vX \mbox{ entails all relations in $\mathcal{C}_{\mathcal{T}}$}\}.
\end{displaymath}
If for each $C_{\pi,i}$,
\begin{displaymath}
	\mathbf{s}_{i} = \{\mathcal{S}_{\pi,i}(\vx_S) : X_{\pi_i} \independent \vX_{[\pi_1:\pi_{i-1}]\setminus S} \mid \vX_S = \vx_S\in\mathcal{C}_{\pi,i}\}
\end{displaymath}
is a partition of $\mathcal{X}_{[\pi_1:\pi_{i-1}]}$, we call the model $\mathcal{M}(\mathcal{T})$ a \emph{CStree model} for the \emph{CStree} $\mathcal{T} = (\pi,\mathbf{s})$ where $\mathbf{s} = \bigcup_{i\in[p]}\mathbf{s}_i$.

A distribution $\vX$ is \emph{Markov} to $\mathcal{T} = (\pi,\mathbf{s})$ if $\vX\in \mathcal{M}(\mathcal{T})$.
The use of the word ``Markov'' suggests a factorization of $P(\vx)$ analogous to that for DAGs in~\eqref{eqn:factorization}.
Indeed, given $i\in [p]$ each outcome $\vx_{\pi_1:\pi_{i-1}}\in \mathcal{S}_{\pi,i}(\vx_S)$ can be mapped to the set of variables $S$ indexing the context $\vx_S$.
Let
\begin{displaymath}
	\cpa_{\mathcal{T}}: \bigcup_{i\in[p]}\mathcal{X}_{[\pi_1:\pi_{i-1}]} \rightarrow \{S: S\subseteq [p]\}
\end{displaymath}
denote this map.
Then $\vX$ is Markov to $\mathcal{T}$ if and only if
\begin{equation*}
	\label{eqn:csfactorization}
	P(\vx) = \prod_{i=1}^p P(x_{\pi_i} | \vx_{\cpa_{\mathcal{T}}(\vx_{\pi_1:\pi_{i-1}})})
\end{equation*}
for all $\vx \in\mathcal{X}$.
The set ${\cpa_{\mathcal{T}}(\vx_{\pi_1:\pi_{i-1}})}$ can be viewed as the \emph{context-specific parents} of $\vx_{\pi_1:\pi_{i-1}}$.
If $\mathbf{s}_{i} = \{\mathcal{S}_{\pi,i}(\vx_{P_i}) : \vx_{P_i}\in\mathcal{X}_{P_i}\}$ for some set $P_i$, for each $i$, then the above factorization reduces to~\eqref{eqn:factorization}, recovering directed acyclic graphical models.
So CStrees are a generalization of directed acyclic graphical models.

The model-defining relations $\mathcal{C}_{\mathcal{T}}$ may be graphically represented by a rooted tree with colored vertices.
Consider a rooted tree $\mathcal{T}$ with vertex set $\{r\}\cup \bigcup_{i\in[2:p]}\mathcal{X}_{[\pi_1:\pi_{i-1}]}$ and edges $\vx_{\pi_1:\pi_{i-1}}\rightarrow \vx_{\pi_1:\pi_{i-1}}x_{\pi_i}$ for all pairs $\vx_{\pi_1:\pi_{i-1}}\in \mathcal{X}_{[\pi_1:\pi_{i-1}]}$ and $x_{\pi_i}\in \mathcal{X}_{\pi_i}$ and $r\rightarrow x_{\pi_1}$ for all $x_{\pi_1}\in\mathcal{X}_{\pi_1}$.
We color the nodes in each set $\mathcal{S}_{\pi,i}(\vx_S)$ the same, using a distinct color for each set.
(When the set $\mathcal{S}_{\pi,i}(\vx_S)$ is a singleton, we abide by the convention of coloring the node white.)
This colored tree then encodes the defining context-specific relations $\mathcal{C}_{\mathcal{T}}$ (and hence the distribution's factorization).

%---FIGURE: Simulation results for p = 6, n = 1000----
\begin{figure}[t]
	\begin{subfigure}[b]{0.45\textwidth}
		\centering
		\begin{tikzpicture}[thick,scale=0.15]
			%---NODES---
			\node[draw, fill=black!0, inner sep=2pt, rounded corners, minimum width=2pt] (w3) at (6,15)  {\scriptsize 1111};
			\node[draw, fill=black!0, inner sep=2pt, rounded corners, minimum width=2pt] (w4) at (6,13.5) {\scriptsize 1110};
			\node[draw, fill=black!0, inner sep=2pt, rounded corners, minimum width=2pt] (w5) at (6,12) {\scriptsize 1101};
			\node[draw, fill=black!0, inner sep=2pt, rounded corners, minimum width=2pt] (w6) at (6,10.5) {\scriptsize 1100};
			\node[draw, fill=black!0, inner sep=2pt, rounded corners, minimum width=2pt] (v3) at (6,9)  {\scriptsize 1011};
			\node[draw, fill=black!0, inner sep=2pt, rounded corners, minimum width=2pt] (v4) at (6,7.5) {\scriptsize 1010};
			\node[draw, fill=black!0, inner sep=2pt, rounded corners, minimum width=2pt] (v5) at (6,6) {\scriptsize 1001};
			\node[draw, fill=black!0, inner sep=2pt, rounded corners, minimum width=2pt] (v6) at (6,4.5) {\scriptsize 1000};
			\node[draw, fill=black!0, inner sep=2pt, rounded corners, minimum width=2pt] (w3i) at (6,3)  {\scriptsize 0111};
			\node[draw, fill=black!0, inner sep=2pt, rounded corners, minimum width=2pt] (w4i) at (6,1.5) {\scriptsize 0110};
			\node[draw, fill=black!0, inner sep=2pt, rounded corners, minimum width=2pt] (w5i) at (6,0) {\scriptsize 0101};
			\node[draw, fill=black!0, inner sep=2pt, rounded corners, minimum width=2pt] (w6i) at (6,-1.5) {\scriptsize 0100};
			\node[draw, fill=black!0, inner sep=2pt, rounded corners, minimum width=2pt] (v3i) at (6,-3)  {\scriptsize 0011};
			\node[draw, fill=black!0, inner sep=2pt, rounded corners, minimum width=2pt] (v4i) at (6,-4.5) {\scriptsize 0010};
			\node[draw, fill=black!0, inner sep=2pt, rounded corners, minimum width=2pt] (v5i) at (6,-6) {\scriptsize 0001};
			\node[draw, fill=black!0, inner sep=2pt, rounded corners, minimum width=2pt] (v6i) at (6,-7.5) {\scriptsize 0000};

			\node[draw, fill=blue!0, inner sep=2pt, rounded corners, minimum width=2pt] (w1) at (-2,14.25) {\scriptsize 111};
			\node[draw, fill=cyan!60, inner sep=2pt, rounded corners, minimum width=2pt] (w2) at (-2,11.25) {\scriptsize 110};
			\node[draw, fill=orange!60, inner sep=2pt, rounded corners, minimum width=2pt] (v1) at (-2,8.25) {\scriptsize 101};
			\node[draw, fill=cyan!60, inner sep=2pt, rounded corners, minimum width=2pt] (v2) at (-2,5.25) {\scriptsize 100};
			\node[draw, fill=red!0, inner sep=2pt, rounded corners, minimum width=2pt] (w1i) at (-2,2.25) {\scriptsize 011};
			\node[draw, fill=cyan!60, inner sep=2pt, rounded corners, minimum width=2pt] (w2i) at (-2,-0.75) {\scriptsize 010};
			\node[draw, fill=orange!60, inner sep=2pt, rounded corners, minimum width=2pt] (v1i) at (-2,-3.75) {\scriptsize 001};
			\node[draw, fill=cyan!60, inner sep=2pt, rounded corners, minimum width=2pt] (v2i) at (-2,-6.75) {\scriptsize 000};

			\node[draw, fill=green!60, inner sep=2pt, rounded corners, minimum width=2pt] (w) at (-8,12.75) {\scriptsize 11};
			\node[draw, fill=cyan!0, inner sep=2pt, rounded corners, minimum width=2pt] (v) at (-8,6.75) {\scriptsize 10};
			\node[draw, fill=green!60, inner sep=2pt, rounded corners, minimum width=2pt] (wi) at (-8,0.75) {\scriptsize 01};
			\node[draw, fill=cyan!0, inner sep=2pt, rounded corners, minimum width=2pt] (vi) at (-8,-5.25) {\scriptsize 00};

			\node[draw, fill=yellow!60, inner sep=2pt, rounded corners, minimum width=2pt] (r) at (-14,9.75) {\scriptsize 1};
			\node[draw, fill=yellow!60, inner sep=2pt, rounded corners, minimum width=2pt] (ri) at (-14,-1.75) {\scriptsize 0};

			\node[draw, fill=black!0, inner sep=2pt, rounded corners, minimum width=2pt] (I) at (-20,3) {\scriptsize r};

			%---EDGES---
			\draw[->]   (I) --    (r) ;
			\draw[->]   (I) --   (ri) ;

			\draw[->]   (r) --   (w) ;
			\draw[->]   (r) --   (v) ;

			\draw[->]   (w) --  (w1) ;
			\draw[->]   (w) --  (w2) ;

			\draw[->]   (w1) --   (w3) ;
			\draw[->]   (w1) --   (w4) ;
			\draw[->]   (w2) --  (w5) ;
			\draw[->]   (w2) --  (w6) ;

			\draw[->]   (v) --  (v1) ;
			\draw[->]   (v) --  (v2) ;

			\draw[->]   (v1) --  (v3) ;
			\draw[->]   (v1) --  (v4) ;
			\draw[->]   (v2) --  (v5) ;
			\draw[->]   (v2) --  (v6) ;

			\draw[->]   (ri) --   (wi) ;
			\draw[->]   (ri) -- (vi) ;

			\draw[->]   (wi) --  (w1i) ;
			\draw[->]   (wi) --  (w2i) ;

			\draw[->]   (w1i) --  (w3i) ;
			\draw[->]   (w1i) -- (w4i) ;
			\draw[->]   (w2i) --  (w5i) ;
			\draw[->]   (w2i) --  (w6i) ;

			\draw[->]   (vi) --  (v1i) ;
			\draw[->]   (vi) --  (v2i) ;

			\draw[->]   (v1i) --  (v3i) ;
			\draw[->]   (v1i) -- (v4i) ;
			\draw[->]   (v2i) -- (v5i) ;
			\draw[->]   (v2i) --  (v6i) ;

			%---LABELS---
			\node at (-17.5,-9) {$X_1$} ;
			\node at (-11.5,-9) {$X_2$} ;
			\node at (-5,-9) {$X_3$} ;
			\node at (2,-9) {$X_4$} ;

		\end{tikzpicture}
		\caption{A CStree $\mathcal{T}$ for variable ordering $\pi = 1234$.}\label{fig:cstree}
	\end{subfigure}
	\hfill
	\begin{subfigure}[b]{0.45\textwidth}
		\centering
		\begin{tikzpicture}[thick,scale=0.4]
			%---NODES---
			\node[circle, draw, fill=black!0, inner sep=1pt, minimum width=1pt] (H1) at (3.25,8) {\large$1$};
			\node[circle, draw, fill=black!0, inner sep=1pt, minimum width=1pt] (B1) at (-2.25,4) {\large$2$};
			\node[circle, draw, fill=black!0, inner sep=1pt, minimum width=1pt] (G1) at (8.25,4) {\large$3$};
			\node[circle, draw, fill=black!0, inner sep=1pt, minimum width=1pt] (B2) at (3.25,0) {\large$4$};

			%---EDGES---
			\draw[->]   (H1) -- node[midway,sloped,above]{${\{1\}}$}(G1) ;
			\draw[->]   (H1) -- node[align=center,below, rotate=-90]{{$\{(0,1),\, (\ast,0)\}$}} (B2) ;
			\draw[->]   (B1) -- node[align=center,below, rotate=-40]{{$\{(\ast,0)\}$}} (B2) ;
			\draw[->]   (G1) -- (B2) ;
			\draw[->]   (B1) -- (G1) ;
		\end{tikzpicture}
		\caption{LDAG representation of the CStree in Figure~\ref{fig:cstree}.}\label{fig:LDAG}
	\end{subfigure}

	\caption{A CStree on four binary variables and its more compact LDAG representation.
		In the LDAG, an edge $i\rightarrow j$ vanishes whenever the outcome $\vx_{\pa_{\mathcal{G}}(j)\setminus i}$ in the edge label is realized.
		The notation $\ast$ indicates that any outcome of the particular variable is included in the edge label.
		Note that the single blue stage in Figure~\ref{fig:cstree} corresponds to all of the context-specific relations captured by the edge labels $(\ast, 0)$ of $1\rightarrow 4$ and the labels of $2\rightarrow 4$.
		This reflects how the single CStree relation $X_4 \independent \vX_{1,2} \mid X_3 = 0$ is encoded in the LDAG, which is defined for pairwise relations, as discussed in Section~\ref{sec:relatedwork}.
		Specifically, this CStree relation implies the pairwise relations $X_4 \independent X_{1} \mid X_2, X_3 = 0$ and $X_4 \independent X_{2} \mid X_1, X_3 = 0$ by basic properties of conditional independence.
	}
	\label{fig:CStreeExample}
\end{figure}
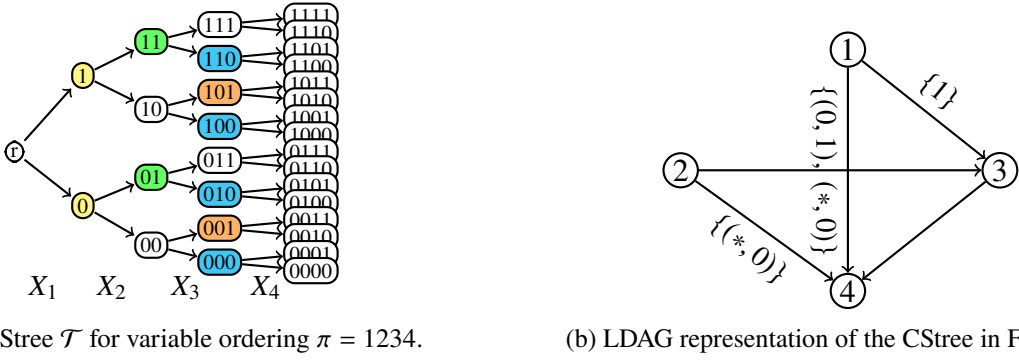

\begin{example}
	\label{ex:cstree}
	The context-specific conditional independence model presented in Example~\ref{ex:CSI} is a CStree model.
	To see this, we let the outcome ``no'' correspond to $0$ and the outcome ``yes'' correspond to $1$.
	Assuming that the covariate $X_1$ is also binary, the model is then \[ \mathcal{C}_{\mathcal{T}} = \{ X_4\independent \vX_{1,2} \mid X_3 = 0,\, X_4 \independent X_1 \mid \vX_{2,3} = (0, 1),\, X_3 \independent X_1 \mid X_2 = 1,\, X_2 \independent X_1 \}, \] for which we take the variable ordering $\pi = 1234$, yielding a pair $(\pi, \mathbf{s})$.
	To see that the sets $\mathbf{s}_i$ partition each $\mathcal{X}_{[\pi_1:\pi_{i-1}]}$, meaning that $(\pi, \mathbf{s})$ is indeed a CStree, we consider the colored tree in Figure~\ref{fig:cstree}.
	The context in the relation $X_3 \independent X_1 \mid X_2 = 1$ is $X_2 = 1$.
	Hence, the nodes $(0,1)$ and $(1,1)$ are the same color (green) in the tree $\mathcal{T}$ in Figure~\ref{fig:cstree}.
	For variable $X_4$, we have relations defined by contexts $X_3 = 0$ and $\vX_{2,3} = (0,1)$, which respectively correspond to the colored sets of nodes $\{(0,0,0),(0,1,0), (1, 0, 0), (1,1,0)\}$ (blue) and $\{(0,0,1),(1,0,1)\}$ (orange) in $\mathcal{T}$.

	Similarly, the context-defining the relation $X_2 \independent X_1$ is the empty context $\vX_{\emptyset} = \vx_{\emptyset}$ where we assume $\vx_\emptyset$ is a subvector of any vector of outcomes.
	Thus, $\mathcal{S}(\vx_{\emptyset}) = \mathcal{X}_{[1]} = \{0,1\}$, and so the two nodes $(0)$ and $(1)$ are colored the same (yellow), representing this relation.
	The white nodes correspond to singleton sets $\mathcal{S}_{\pi, i}(\vx_S)$ since the corresponding variables were not assumed independent of any preceding variables in these contexts $\vx_S$.
	Since none of these colored sets overlap in Figure~\ref{fig:cstree}, the sets $\mathbf{s}_i$ do indeed partition each $\mathcal{X}_{[\pi_1:\pi_{i-1}]}$.
	So $\mathcal{T} = (\pi, \mathbf{s})$ is a CStree.
\end{example}

The tree $\mathcal{T}$ in Figure~\ref{fig:cstree} is an example of a staged tree representation of a context-specific model, and it shows why staged trees are difficult to interpret even for very few variables.
As the number of variables and state spaces for each variable grow, we simply cannot draw or interpret such a graphical representation.

Following the staged tree literature, we call the set $\mathcal{S}_{\pi,i}(\vx_S)$ a \emph{stage}, $\vx_S$ its \emph{stage-defining context} and $S$ its set of \emph{context variables}.
Each stage corresponds to a set of colored nodes in $\mathcal{T}$.
We call the set of outcomes $\mathcal{X}_{[i]}$ \emph{level $i$} of $\mathcal{T} = (\pi,\mathbf{s})$ and $\mathbf{s}_i$ a \emph{staging of level $i$}.
The set $\mathbf{s}$ is a \emph{staging} of the tree.
The stage $s = \mathcal{S}_{\pi,i}(\vx_S)\in\mathbf{s}_i$ corresponds to a conditional probability $P(X_{\pi_i} | \vx_S)$ used in the factorization~\eqref{eqn:csfactorization}.
We parametrize this conditional distribution with $\mathbf{\theta}_{\pi_i,s} = [\theta_{\pi_i,s,1},\ldots,\theta_{\pi_i,s,d_{\pi_i}}]$, where $\theta_{\pi_i,s,t}$ is the probability that $X_{\pi_i} = t$ given $\vx_S$.
We let $\theta_{\pi,\mathbf{s}_i}$ denote the concatenation of the vectors $\mathbf{\theta}_{\pi_i,s}$ over all stages $s\in \mathbf{s}_{i}$, and similarly we let $\theta_{\pi,\mathbf{s}} = [\theta_{\pi,\mathbf{s}_1},\ldots, \theta_{\pi,\mathbf{s}_p}]$.

While staged trees are difficult to work with, repeated application of the \emph{decomposition property} \citep{corander2019logical} to the relations in $\mathcal{C}_{\mathcal{T}}$ shows that any distribution Markov to a CStree is also Markov to an LDAG.
Thus, a CStree also admits a much more compact LDAG representation, depicted for our example in Figure~\ref{fig:LDAG}.

Generally, an LDAG is a pair $(\mathcal{G}, \mathcal{L})$ where $\mathcal{G} = ([p], E)$ is a directed acyclic and $\mathcal{L} = \{L_{i,j} : i\rightarrow j \in E\}$ is a collection of sets, called labels, with one for each edge of $\mathcal{G}$.
The label $L_{i,j}$ is a set of outcomes of $\vX_{\pa_{\mathcal{G}}(j)\setminus i}$.
A distribution $\vX$ is Markov to $(\mathcal{G}, \mathcal{L})$ if $\vX$ is Markov to $\mathcal{G}$ and $X_j \independent X_i \mid \vX_{\pa_{\mathcal{G}}(j)\setminus i} = \vx_{\pa_{\mathcal{G}}(j)\setminus i}$ for all $\vx_{\pa_{\mathcal{G}}(j)\setminus i}\in L_{i,j}$, for all $i\rightarrow j \in E$.
Hence, LDAGs are defined via pairwise context-specific relations, unlike CStrees which are defined according to a context-specific generalization of the factorization definition~\eqref{eqn:factorization} of a DAG model.
Details on how to construct an LDAG for a CStree model are given in the Supplement \citep{RMS2024supplement}.

Note that since the pairs of nodes $\{(0,0), (1,0)\}$ and $\{(0,1,0),(0,0,0)\}$ are not assigned a color in $\mathcal{T}$ in Figure~\ref{fig:cstree}, we do not have that the relations $X_3 \independent X_1 | X_2 = 0$ and $X_4 \independent X_2 | X_{\{1,3\}} = (0,0)$.
This implies that the distributions in $\mathcal{M}(\mathcal{T})$ need not entail the relations $X_3 \independent X_1 |X_2$ and $X_4 \independent X_2 | X_{\{1,3\}}$.
In particular, the parents for $X_3$ and $X_4$ in a (minimal) directed acyclic graph representation of this model are, respectively, $\{X_1,X_2\}$ and $\{X_1,X_2,X_3\}$.
In other words, a minimal I-MAP of the CStree model $\mathcal{M}(\mathcal{T})$ is the graph $\G$ in the LDAG $(\mathcal{G}, \mathcal{L})$ depicted in Figure~\ref{fig:LDAG}.

\begin{remark}
	\label{rem: model equivalence}
	Analogous to directed acyclic graphical models, it is possible that $\mathcal{M}(\mathcal{T}) = \mathcal{M}(\mathcal{T}^\prime)$ for two CStrees $\mathcal{T}\neq \mathcal{T}^\prime$, in which case we say that $\mathcal{T}$ and $\mathcal{T}^\prime$ are \emph{Markov equivalent} \citep{duarte2021representation}.
	In particular, without additional assumptions on the data-generating distribution, one cannot distinguish from a random sample whether the data was generated according to the structure of $\mathcal{T}$ or $\mathcal{T}^\prime$.
	This nonidentifiability issue must be taken into consideration in structure learning algorithms.
	Here, we use estimators that score equally on Markov equivalent CStree models in order to avoid spurious preferences in model selection (see Section~\ref{subsec:score}).
\end{remark}

\section{Scoring CStrees and a Context-specific Sparsity Constraint}\label{subsec:score}
We use a score function for staged trees called the CS-BDeu score of \citet{hughes2022bayesian}.
The score uses a Dirichlet prior on the model parameters which ensures that Markov equivalent CStrees are scored equally, as discussed in Remark~\ref{rem: model equivalence}.

Including the parameters in the factorization~\eqref{eqn:csfactorization}, we obtain
\begin{displaymath}
	P(\vx |\mathbf{\theta}_{\pi,\mathbf{s}}, \pi,\mathbf{s}) = \prod_{i=1}^pP(x_{\pi_{i}}|\vx_{\cpa_{\mathcal{T}}(\vx_{\pi_1:\pi_{i-1}})} ,\mathbf{\theta}_{\pi,\mathbf{s}}, \pi,\mathbf{s}).
\end{displaymath}
Assuming independent parameters $\theta_{\pi,s}$ for each stage, we define for node $i$ and stages enumerated as $1,\dots,q_i$,
\begin{displaymath}
	P(\theta_{\pi,\mathbf{s}} | \pi, \mathbf{s}) = \prod_{i=1}^p\prod_{s=1}^{q_i}P(\theta_{\pi_i,s} | \pi,\mathbf{s}).
\end{displaymath}
As a prior, we take each $\theta_{\pi_i,s} | \pi, \mathbf{s}$ to be Dirichlet so that
\begin{equation*}
	\begin{split}
		P(\theta_{\pi_i,s} | \pi,\mathbf{s}) = \mathrm{Dir}(\theta_{\pi_i,s} | \alpha_{{\pi_i}s1},\ldots,\alpha_{{\pi_i}sd_{\pi_i}})\propto \prod_{k=1}^{d_{\pi_i}}\theta_{\pi_isk}^{\alpha_{isk}-1}.
	\end{split}
\end{equation*}
The resulting marginal likelihood function (see the Supplement \citep{RMS2024supplement} for details) is
\begin{equation}
	\label{eq:marglik}
	P(\vx | \pi, \mathbf{s}) = \prod_{\substack{i=1 \\ s= 1}}^{p,q_i}\frac{\Gamma(\alpha_{\pi_is})}{\Gamma(\alpha_{\pi_is} + N_{\pi_is})}\prod_{k=1}^{d_{\pi_i}}\frac{\Gamma(\alpha_{\pi_isk} + N_{\pi_isk})}{\Gamma(\alpha_{\pi_isk})},
\end{equation}
where $N_{isk}$ denotes the number of data points $\vy$ in which $y_i = k$ and $\vy_S$ is the stage-defining context of $s$.
We also let $N_{is} = \sum_{k=1}^{d_{\pi_i}} N_{isk}$ and $\alpha_{is} = \sum_{k=1}^{d_{\pi_i}}\alpha_{isk}$.
Notice that this likelihood is \emph{decomposable} with respect to the ordering $\pi$ as it is a product of terms over $i = 1,\ldots,p$.

Similarly, we work with a decomposable prior on $(\pi,\mathbf{s})$:
\begin{displaymath}
	P(\pi, \mathbf{s}) = P(\mathbf{s} | \pi)P(\pi) = \prod_{i=1}^pP(\mathbf{s}_i| \pi_{1:i})P(\pi_{1:i}),
\end{displaymath}
where we take $P(\pi_{1:i}) = 1/i$ and $P(\mathbf{s}_i | \pi_{1:i}) = 1/|\mathcal{S}|$, with $\mathcal{S}$ denoting the set of allowed stagings of level $i$.

Recall from Section~\ref{sec:cstrees} that a CStree $\mathcal{T}$ is simply a pair $(\pi, \mathbf{s})$.
So the above prior is a prior over all CStrees for the variables $X_1,\ldots,X_p$.
Given such a decomposable prior, we recover the posterior of a CStree, which is also decomposable:
\begin{equation}
	\label{eqn:jointpost}
	\begin{split}
		P(\pi, \mathbf{s} | \vx) \propto P(\vx | \pi, \mathbf{s}) P(\pi, \mathbf{s}) = \prod_{i=1}^pP(\vx_{\pi_1:\pi_{i}} | \pi_{1:i}, \mathbf{s}_i)P(\mathbf{s}_i | \pi_{1:i})P(\pi_{1:i}).
	\end{split}
\end{equation}

We will use the score in two phases of our structure learning algorithm: first to estimate an optimal variable ordering and then to identify the stages $\mathbf{s}_i$ for each $i$.
For these purposes, we require an ordering score, which we take to be the unnormalized marginal order posterior.
\begin{equation}
	\label{eqn:orderposterior}
	\begin{split}
		\tilde{P}(\pi | \vx)
		 & = \sum_{\mathbf{s}}
		P(\vx | \pi, \mathbf{s}) P(\pi, \mathbf{s}), \\ &= \sum_{\mathbf{s}} \prod_{i=1}^pP(\vx_{\pi_1:\pi_{i}} | \pi_{1:i}, \mathbf{s}_i)P(\mathbf{s}_i | \pi_{1:i})P(\pi_{1:i}),\\ & = \prod_{i=1}^p \sum_{\mathbf{s}_i} P(\vx_{\pi_1:\pi_{i}} | \pi_{1:i}, \mathbf{s}_i)P(\mathbf{s}_i | \pi_{1:i})P(\pi_{1:i}).
	\end{split}
\end{equation}
Since, for stage $s = \mathcal{S}(\vx_S)$, the \emph{context marginal likelihood}
\begin{equation}
	\label{eqn:contextmarginallikelihood} z_{i,\vx_S} = \frac{\Gamma(\alpha_{is})}{\Gamma(\alpha_{is} + N_{is})}\prod_{k=1}^{d_{i}}\frac{\Gamma(\alpha_{isk} + N_{isk})}{\Gamma(\alpha_{isk})}
\end{equation}
depends only on the variable ${i}$ and the stage-defining context $\vx_S$ for stage $s$, these values may be precomputed and stored.
This may be done efficiently when the possible stage-defining contexts $\vx_S$ are limited according to the following novel context-specific sparsity constraint.
For a CStree $\mathcal{T} = (\pi, \mathbf{s})$, let $\ms(\mathbf{s}_i)$ denote the maximum size of a set $S$ for which there is a set $\mathcal{S}_{\pi, i}(\vx_S)\in \mathbf{s}_i$.

%---ASSUMPTION---
\begin{assumption}
	\label{ass:CS}
	Fix a positive integer $\beta$.
	The data-generating distribution $\vX$ is Markov to a CStree $\mathcal{T} = (\pi, \mathbf{s})$ such that $\ms(\mathbf{s}_i) \leq \beta$ for all $i$.
\end{assumption}
Accounting for Assumption~\ref{ass:CS}, we need only compute the $z_{i,\vx_S}$ for each pair $(i,\vx_S)$ where $|S|\leq \beta$ and $\vx_S$ defines a stage in some CStree.
Doing so requires enumerating all stagings of level $i$ of a CStree for the chosen $\beta$, which, after some derivations in the Supplement \citep{RMS2024supplement}, is possible for small $\beta$.
\begin{theorem}
	\label{thm:enumeratingstagings}
	There are $1 - \binom{i}{2} + \sum_{k=1}^ii^{d_k}$ stagings $\mathbf{s}_i$ of level $i$ in which each stage has at most two context variables; that is, stagings such that $\ms(\mathbf{s}_i) \leq 2$.
\end{theorem}
The special case of Theorem~\ref{thm:enumeratingstagings} in which $d_k = 2$ for all $k\in[p]$ solves a problem posed by \citet{alon2023partitioning} (see Corollary~1 in the Supplement \citep{RMS2024supplement}).
The most general version of their problem corresponds to results as in Theorem~\ref{thm:enumeratingstagings} for arbitrary $\beta \geq 1$ and $d_k = 2$.
In particular, extending the results of Theorem~\ref{thm:enumeratingstagings} to higher values of $\beta$ will solve increasingly harder cases of their challenging combinatorial problem.
Regarding inference, it will allow for the methods developed here to be applied to increasingly dense systems (e.g., a relaxation of the bound $\beta=2$).

\begin{remark}
	\label{rem: not as sparse as you think}
	While the sparsity bound $\beta = 2$ may seem restrictive it is worth noting two facts.
	First, a CStree $\mathcal{T}$ satisfying Assumption~\ref{ass:CS} for $\beta = 2$ may have more than two parents per node in its LDAG representation.
	In other words, the context-specific sparsity bound $\beta = 2$, while analogous to bounding the size of parent sets in a directed acyclic graph, in fact allows for the directed acyclic graph $\mathcal{G}$ in the LDAG representation of $\mathcal{T}$ to be rather dense.
	Further details are in Remark~3 in the Supplement \citep{RMS2024supplement}.

	Second, as noted in the introduction, a sparse representation of the data amounts to a reduced number of parameters in the resulting representation of the conditional probability table which is, in fact, advantageous when it comes to performing predictive inference tasks with the model \citep{koller2009probabilistic}.
	In particular, most classical optimization-based directed acyclic graph learning algorithms are designed to prefer sparser models and empirically only exhibit very accurate results in such sparse settings.
\end{remark}

Imposing Assumption~\ref{ass:CS} for a fixed $\beta$ need not prevent one from incorporating a second, more classical, sparsity assumption, if it is desired.

%---ASSUMPTION: DAG Sparsity---
\begin{assumption}[Classic DAG sparsity]
	\label{ass:DAG}
	Let $\alpha$ be a positive integer.
	The distribution $\vX$ is Markov to a CStree $\mathcal{T}$ whose LDAG representation $(\mathcal{G}, \mathcal{L})$ satisfies $|\pa_{\mathcal{G}}(i)|\leq \alpha$ for all $i$.
\end{assumption}

Assumption~\ref{ass:DAG} need not be specific to CStrees, having already been used in staged tree learning \citep{leonelli2022highly}, LDAG learning \citep{hyttinen2018structure, pensar2015labeled} and even DAG learning \citep{cussens2012bayesian}.
However, Assumption~\ref{ass:CS} is specific to the definition of CStrees, and critical to the efficiency achieved by our proposed method.

To incorporate both Assumption~\ref{ass:DAG} and Assumption~\ref{ass:CS}, we may constrain the staging $\mathbf{s}_i$ such that every $\mathcal{S}(\vx_S)\in\mathbf{s}_i$ satisfies $S\subseteq K_i$ for some $K_i\subseteq[p]$ and $|S|\leq \beta$.
The sets $K_i$ correspond to the possible parents of $i$ in the graph $\mathcal{G}$ in the LDAG representation of $\mathcal{T}$.
Let $\mathcal{S}_{L_,\beta}$ denote this set of stagings for any $L\subseteq K_i$.
Then we may precompute the \emph{local ordering scores}
\begin{equation}
	\label{eqn:loscores} \textrm{los}(i,L; \vx) = \sum_{\mathbf{s}_i\in\mathcal{S}_{L,\beta}}\frac{1}{i|\mathcal{S}_{L,\beta}|}\prod_{\mathcal{S}(\vx_S)\in\mathbf{s}_i}z_{i,\vx_S}.
\end{equation}
Given a variable ordering $\pi$ and sets $K_1,\ldots, K_p$, the relevant sets $L$ are $L_i = K_i \cap [\pi_1:\pi_{i-1}]$.
Hence, we may use the precomputed local ordering scores to compute $\tilde P(\pi| \vx)$.

This is the most computationally intensive step of the proposed method.
Details are provided in the Supplement \citep{RMS2024supplement}, including a proof of the following complexity result:
\begin{theorem}
	\label{thm:loscorecomp}
	Let $\beta > 0$, $d = \textrm{max}_{i}(d_i)$ and $K_1,\ldots, K_p\subseteq[p]$.
	The local ordering scores may be computed in $\mathcal O(p2^{|K|}|\mathcal S_{K,\beta}|d^\beta)$, where $K = \textrm{argmax}_{\{K_i : i\in[p]\}}|K_i|$.
\end{theorem}

Assumption~\ref{ass:DAG} then corresponds to the bound $|K|\leq \alpha$.
Since the complexity in Theorem~\ref{thm:loscorecomp} is exponential in both the context-specific sparsity constraint $\beta$ and the classical directed acyclic graph sparsity constraint $|K|\leq \alpha$, controlling $\alpha$ and $\beta$ naturally allows for more efficient and scalable structure learning methods.

\begin{algorithm}
	\caption{(CSlearn) An algorithm for estimating a CStree.}\label{alg:cstreelearn}
	\begin{tabbing}
		\qquad {\bf Require:}
		A random sample $\mathbf{D} = \{\vx^j\}_{j=1}^n$ where $\vx^j = (x_1^j,\ldots,x_p^j)$.
		\\
		\qquad {\bf Require:}
		A context size bound $\beta$\\

		\qquad \enspace 1:\,\, $G = ([p],E)$ $\gets$ CPDAG of a DAG estimate for $\mathbf{D}$ \qquad \qquad $\#$ Option: Apply bound $\alpha$\\ \qquad \enspace 2:\,\, {\bf For} $i \gets 1$ to $p$ \\ \qquad \enspace 3:\,\, \qquad $K_i$ $\gets$ $\{j\in[p] : i - j \in E \mbox{ or } j\rightarrow i\in E\}$\\ \qquad \enspace 4:\,\, Compute marginal context scores for $\beta$ and $K_1,\ldots, K_p$\\ \qquad \enspace 5:\,\, Compute local order scores for $\beta$ and $K_1,\ldots, K_p$\\ \qquad \enspace 6:\,\, Generate samples $\pi^1, \ldots, \pi^t$ from $P(\pi \mid \vx)$ using Gibbs sampler described in Section~\ref{subsec:stochastic}\\ \qquad \enspace 7:\,\, $\pi^\ast$ $\gets \mathrm{argmax}_{\pi^i:i \in [t]} P(\pi^i |\vx)$\\ \qquad \enspace 8:\,\, $\mathbf{s}^\ast$ $\gets [\,\,\,]$\\ \qquad \enspace 9:\,\, {\bf For} $i \gets 1$ to $p$\\ \qquad \enspace 10: \qquad $L$ $\gets$ $\{j\in K_i : \mbox{$j$ precedes $i$ in $\pi^\ast$}\}$\\ \qquad \enspace 11: \qquad $\mathbf{s}_i^\ast$ $\gets \mathrm{argmax}_{\mathbf{s}_i\in\mathcal{S}_{L,\beta}} P(\mathbf{s}_i |\vx, \pi_{1:i}^\ast)$\\ \qquad \enspace 12: \qquad Append $\mathbf{s}_i^\ast$ to $\mathbf{s}^\ast$\\ \qquad \enspace 13: {\bf Return} $(\pi^\ast, \mathbf{s}^\ast)$
	\end{tabbing}
\end{algorithm}

\section{A Structure Learning Algorithm}\label{subsec:alg} The proposed algorithm, presented in Algorithm~\ref{alg:cstreelearn} proceeds in three phases.
It starts with a variable selection phase, followed by a stochastic search, and finally an exact optimization phase.

\subsection{Possible Context-Variable Selection Phase.}
\label{subsec:constraint}
The first phase of Algorithm~\ref{alg:cstreelearn} is optional, serving as an opportunity to incorporate classic graph sparsity (Assumption~\ref{ass:DAG}) if it is desired.
In this phase, we limit the possible indices $i$ that may appear in the set $S$ of a stage-defining context $\vx_S$.
This is Step~1 of Algorithm~\ref{alg:cstreelearn}, which estimates an \emph{essential graph} representation (also known as a CPDAG) of the distribution $\vX$.
In Steps 2 and 3, for each node $i$, the nodes that are possible parents of $i$ according to the essential graph are collected into a set $K_i$.
The CStrees $\mathcal{T}= (\pi, \mathbf{s})$ considered in the coming phases are those that, for all $i$, satisfy $S\subseteq K_i$ for all $\mathcal{S}_{\pi, i}(\vx_S)\in \mathbf{s}_i$.
In the language of Assumption~\ref{ass:DAG}, including this phase imposes the sparsity constraint $\alpha = \max_i|K_i|$.

This phase is the optionally constraint-based phase, since one may choose to use any constraint-based or optimization-based directed acyclic graph learning algorithm.
However, essential graph estimation may be done with any directed acyclic graph learning algorithm, although the choice of algorithm may impact efficiency of the overall method.
In our experiments in Section~\ref{sec:learningex}, we demonstrate the performance of CSlearn where we use the classic constraint-based PC algorithm \citep{spirtes1991algorithm}, as well as one of the more recent and currently best performing hybrid algorithms, known as GRaSP \citep{lam2022greedy}.
One could alternatively take $G$ to be an estimated skeleton of an essential graph and $|K_i|$ to be the neighbors of node $i$ in $G$.
This approach may allow for the later phases of Algorithm~\ref{alg:cstreelearn} to correct any mis-specified arrow directions learned by the selected directed acyclic graph structure learning method.

Steps~4 and~5 are the most computationally intensive steps, precomputing the local order scores.
Larger cardinalities $|K_i|$ or larger $\beta$ will naturally increase runtime (see Theorem~\ref{thm:loscorecomp}).
In our implementation, and our experiments in Section~\ref{sec:learningex}, we impose no apriori bounds on $|K_i|$ and default set $\beta = 2$.
Reasonably larger $\beta$ likely still run fast, but require generalizing Theorem~\ref{thm:enumeratingstagings}.

\subsection{Markov Chain Monte Carlo Phase.}\label{subsec:stochastic}
A stochastic optimization phase is then used to estimate an optimal variable ordering $\pi^\ast$.
This is Steps~6 and~7 in Algorithm~\ref{alg:cstreelearn}.
We take the approach of \citet{friedman2003being, kuipers2022efficient} for directed acyclic graph structure learning and generate a Markov chain on the ordering space.
In particular, we use the Markov chain Monte-Carlo sampler based on the relocation move introduced by \citet{kuipers2022efficient}.
At each stage in the chain, we pick uniformly at random a node $i$ in the order and calculate $\tilde{P}(\pi | \vx)$ when $i$ is placed in every position in the order $\pi$, including where it started.
We then relocate $i$ to any of these positions at random with probability proportional to the calculated scores.
\citet{kuipers2022efficient} show that the chain will converge to the target ordering distribution, and the decomposabilty property enables the relocation procedure to be done in $\mathcal O(p)$ time by using pairwise swapping.

Note that in the case of directed acyclic graphical models, the joint prior over orderings and graphs is specified hierarchically as $P(\mathbf \pi|G)P(G)$.
However, $P(\mathbf \pi|G)$ is typically ignored since it is hard to compute, leading to biased estimates of the order posterior.
To address this issue \cite{kuipers2017partition} consider operating on the space of node partitions.
For CStree models, the pair $(\mathbf \pi, \mathbf s)$ uniquely determines the CStree.
Hence, \eqref{eqn:orderposterior} is the true unnormalized ordering posterior and the Markov chain is guaranteed to have $P(\mathbf \pi | \vx)$ as the stationary distribution.
In particular, our method provides quantification of the uncertainty in the order estimate via this marginal posterior.
In the case of directed acyclic graphs, computing the score of a variable ordering requires summing over all graphs that are consistent with the ordering.
\citet{friedman2003being, kuipers2022efficient} note that this can take exponential-time, despite not needing to account for context-specific constraints.
Hence, it is typical to impose sparsity constraints to allow for reasonable computation time.
For directed acyclic graphs, the sparsity constraints bound the number of possible parents for node $i$ (e.g., $\alpha$ in Assumption~\ref{ass:DAG}).

For CStrees, we introduce the novel context-specific sparsity constraint $\ms(\mathbf{s}_i)\leq \beta$ in Assumption~\ref{ass:CS} to play the analogous role.
Setting $\beta = 1$ in Algorithm~\ref{alg:cstreelearn} learns models that can be interpreted as a context-specific generalization of the directed tree models studied by \citet{jakobsen2022structure}.
Taking $\beta = 2$ (default in our implementation) allows for denser context-specific models (see Remark~\ref{rem: not as sparse as you think}) while still allowing for a high level of scalability.
As noted in Remark~\ref{rem: not as sparse as you think}, the choice of $\beta = 1$ or $\beta = 2$ does not mean that the learned CStree will have LDAG representation $(\mathcal{G}, \mathcal{L})$ where $\mathcal{G}$ satisfies Assumption~\ref{ass:DAG} with $\alpha = 1$ or $\alpha = 2$, respectively.
Indeed, the context-specific nature of the bound $\beta$, allows for denser $\mathcal{G}$.
Following future work generalizing Theorem~\ref{thm:enumeratingstagings}, larger $\beta$ can readily be implemented in our \href{https://cstrees.readthedocs.io/en/latest/index.html}{package}, allowing for even denser models.

\subsection{Exact Optimization Phase.}\label{subsec:exact}
The exact optimization phase estimates the staging $\mathbf{s}$ defining the CStree in Steps~8-12 of Algorithm~\ref{alg:cstreelearn}.
Here we use the context marginal likelihoods, computed and stored in step~4, to exactly identify the optimal staging $\mathbf{s}_i^\ast$ for each $i\in[p]$.
Namely, for each staging of level $i$, we compute the products $P(\vx_{\pi_1^\ast:\pi_{i}^\ast} | \pi_{1:i}^\ast, \mathbf{s}_i)P(\mathbf{s}_i | \pi_{1:i}^\ast)P(\pi_{1:i}^\ast)$ in~\eqref{eqn:jointpost}.
This can be done independently for each $i\in[p]$, and the resulting product over all $i$ is proportional to the conditional posterior $P(\mathbf{s} | \pi^\ast, \vx)$.
In particular, our resulting staging estimate $\mathbf{s}^\ast$ is the maximum a posteriori for this conditional posterior.

Algorithm~\ref{alg:cstreelearn} returns the pair $(\pi^\ast,\mathbf{s}^\ast)$ which is a CStree.
As described in Section~\ref{sec:cstrees}, one can obtain whichever representation of the CStree they desire from $(\pi^\ast,\mathbf{s}^\ast)$; for instance, the staged-tree $\mathcal{T}$ in Figure~\ref{fig:cstree} or the compact LDAG in Figure~\ref{fig:LDAG}.

\section{Experiments}\label{sec:learningex}

\subsection{Accuracy Evaluation}\label{subsec:accuracy}
We empirically evaluated the proposed structure learning algorithm (Algorithm~\ref{alg:cstreelearn}) in terms of accuracy and scalability on synthetic data.
We performed two sets of experiments: one to empirically evaluate the accuracy of Algorithm~\ref{alg:cstreelearn} for different choices of parameter estimators and methods for Phase~1 (i.e., methods for selecting sets of possible context variables, as described in Subsection~\ref{subsec:constraint}), and a second to evaluate how the algorithm scales to large systems ($p \gg 100$).
In all experiments, the Monte-Carlo step of CSlearn (Algorithm~\ref{alg:cstreelearn}) was run with $5000$ iterations.

In the first round of experiments, we generate random CStree models and draw random samples from each model.
Given the sample, Algorithm~\ref{alg:cstreelearn} is tasked with estimating not only the model structure $\mathcal{T} = (\pi,\mathbf{s})$ but also the model parameters $\theta_{\pi,\mathbf{s}}$.
For the parameter estimation step, the current implementation of CSlearn can return either the maximum a posteriori $\hat\theta_{\pi,\mathbf{s}}$ of $P( \theta_{\pi,\mathbf{s}} | \hat{\pi}, \hat{\mathbf{s}}, \vx)$, or the maximum likelihood estimate $\theta_{\pi,\mathbf{s}}^\ast$ for parameters $\theta_{\pi,\mathbf{s}}$ of the estimated CStree $(\hat{\pi}, \hat{\mathbf{s}})$.
We denote a parameterized CStree model as $\mathcal{T} = (\pi,\mathbf{s}, \theta_{\pi,\mathbf{s}})$.

To evaluate the performance of Algorithm~\ref{alg:cstreelearn} at this task, we compute the KL-divergence of each learned CStree model $\hat{\mathcal{T}} = (\hat{\pi},\hat{\mathbf{s}}, \hat{\theta}_{\pi,\mathbf{s}})$ and $\hat{\mathcal{T}} = (\hat{\pi},\hat{\mathbf{s}}, {\theta}_{\pi,\mathbf{s}}^\ast)$, one for each parameter estimator, from the data-generating model $\mathcal{T} = (\pi,\mathbf{s}, \theta_{\pi,\mathbf{s}})$.
Computing KL-divergence requires computing the probability of each outcome in the joint state space $\mathcal{X}$ of $(X_1,\ldots, X_p)$, and we must do this computation for both the estimated distribution and the true distribution.
Hence, this evaluation method can be time-consuming (not due to the structure learning, but instead due to the computation of the KL-divergences).
Thus, for these experiments, we limit our analysis to models where $X_1,\ldots, X_p$ are all binary with $2^p$ elements in the joint state space, and $p \in\{5,7,10,15,20\}$.

For each choice of $p$, $10$ random ground truth CStree models $\mathcal{T} = (\pi,\mathbf{s}, \theta_{\pi,\mathbf{s}})$ are generated using the random CStree generator included in our package.
Since Algorithm~\ref{alg:cstreelearn} estimates a CStree model in which all stages $\mathcal{S}(\vx_S)$ have stage-defining contexts $S$ satisfying $|S|\leq 2$ (i.e., $\beta = 2$), we generate only CStree models that also fulfill this condition.

\begin{remark}
	\label{rem:choice of KL}
	KL-divergence is used as an approximate measure of accuracy for two reasons.
	First, we would like to offer a reasonable comparison against the LDAG learning methods of \citet{hyttinen2018structure}.
	As no implementation of their methods was linked in their paper, we present KL-divergences as they did for their experimental results on exact optimization methods.

	Second, as noted in Remark~\ref{rem: model equivalence}, while the data may be generated from the CStree $\mathcal{T}$, our algorithm may return a CStree $\mathcal{T}^\prime$ satisfying $\mathcal{M}(\mathcal{T}) = \mathcal{M}(\mathcal{T}^\prime)$; e.g. a (Markov) equivalently accurate representation of the data-generating distribution.
	For directed acyclic graphical models, there exist characterizations of Markov equivalence that are efficiently checked \citep{verma1991equivalence}, allowing one to verify if the learned model is equivalent to the data-generating structure.
	The only known characterizations of Markov equivalence of CStrees \citep{duarte2021representation} are computationally infeasible for even small graphs, meaning that structural metrics for checking accuracy in learning (e.g., structural Hamming distance and comparison of stage-defining contexts) are not feasible metrics of evaluation for this problem.
	Specifically, such structural metrics may present misleading results while also not offering any informative comparison with the LDAG methods of \cite{hyttinen2018structure}.
	On the other hand, when the KL-divergence is $0$, the learned distribution is necessarily contained in the model $\mathcal{M}(\mathcal{T})$ for the data-generating CStree $\mathcal{T}$.
\end{remark}

To get a sense of how the choice of parameter estimation impacts model accuracy, we ran these experiments for sample sizes $n \in \{250, 500, 1000, 10000\}$ and computed the KL-divergence $\kl{\hat{\mathcal{T}}}{\mathcal{T}}$ for each fitted CStree $\hat{\mathcal{T}} = (\hat{\pi},\hat{\mathbf{s}}, \hat{\theta}_{\pi,\mathbf{s}})$ and $\hat{\mathcal{T}} = (\hat{\pi},\hat{\mathbf{s}}, {\theta}_{\pi,\mathbf{s}}^\ast)$.
To evaluate the impact of the choice of directed acyclic graph structure learning algorithm used in Phase~1, we performed these experiments using both the PC algorithm \citep{spirtes1991algorithm} and GRaSP \citep{lam2022greedy} for this phase.
The PC algorithm is a classic constraint-based algorithm, whereas GRaSP is one of the top performing algorithms to-date according to the open-source structure learning algorithm benchmarking software \texttt{Benchpress} \citep{rios2021benchpress}.

\begin{figure}
	\begin{subfigure}[b]{0.475\textwidth}
		\centering
		\includegraphics[width=\linewidth]{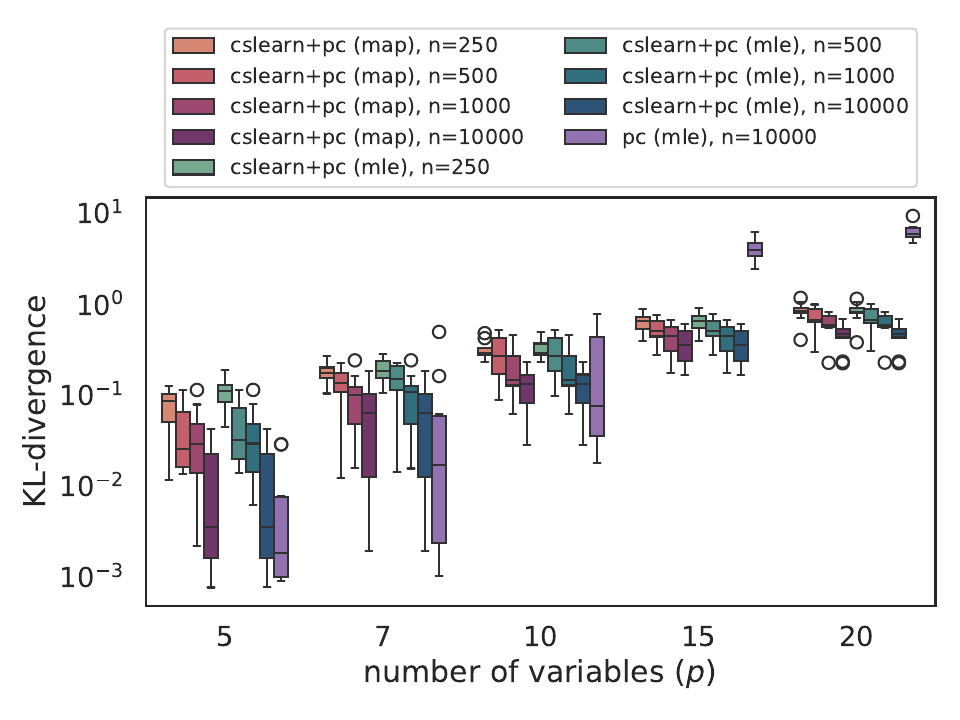}
		\caption{Phase 1 using PC}
		\label{fig:kl-div-2a}
	\end{subfigure}
	\hfill
	\begin{subfigure}[b]{0.475\textwidth}
		\centering
		\includegraphics[width=\linewidth]{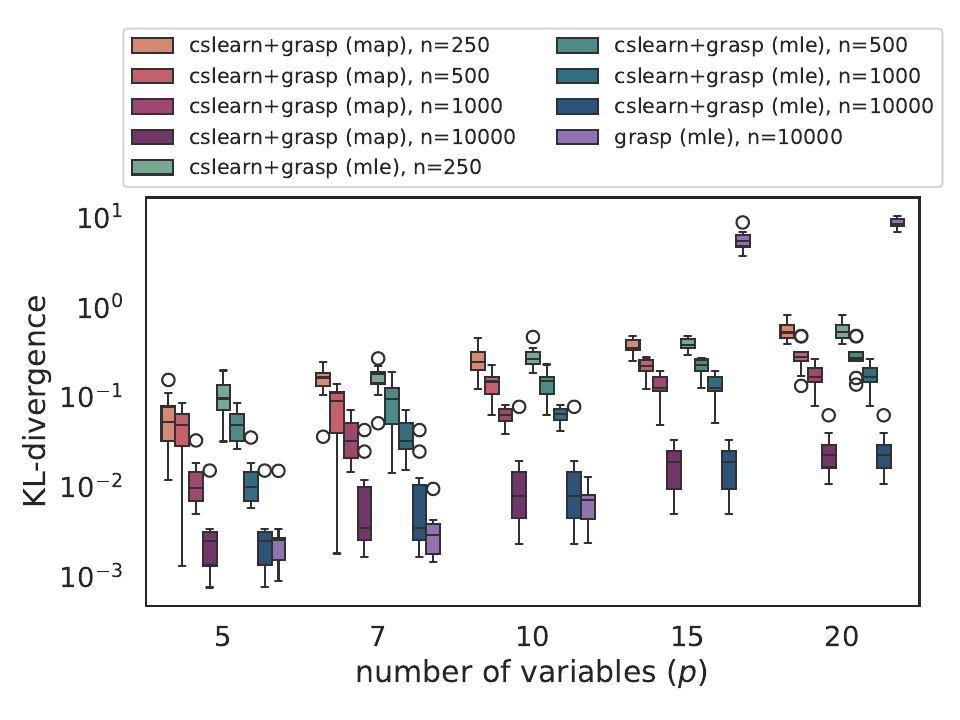}
		\caption{Phase 1 using GRaSP}
		\label{fig:kl-div-2b}
	\end{subfigure}
	\caption{Accuracy of CSlearn for different choices of Phase 1 methods and parameter estimators.
		Plots present results on a semilog scale.
	}
	\label{fig:kl-div}
\end{figure}

The results are presented in Figures~\ref{fig:kl-div-2a} and~\ref{fig:kl-div-2b}, respectively.
We note that the maximum a posteriori and maximum likelihood estimators appear to perform approximately the same across all choices of $n$ and $p$.
We also include the accuracy of the directed acyclic graphical model learned in Phase~1 to explicitly show the performance of a model where additional steps are taken to include context-specific observations.
For these models, we compute only the KL-divergence for the model estimated from $n = 10000$ samples, so as to give the best possible performing directed acyclic graph in the comparisons.
For both PC and GRaSP, KL-divergence increases rapidly when the number of variables increases past $10$, even for the large sample size.
For CSlearn, and all sample sizes, the KL-divergence remains close to $0$ even as the number of variables increases.

In relation to other context-specific learning methods, we can focus on the case of $p = 10$ and $n = 250$ in Figures~\ref{fig:kl-div-2a} and~\ref{fig:kl-div-2b}, where we see a median KL-divergence of approximately $0.15$.
This may be compared with the experiments in \citep[Figure 3]{hyttinen2018structure}, where they use exact optimization techniques to estimate LDAGs on $p = 10$ binary variables with a sample size of $n=200$ (see Remark~\ref{rem:choice of KL}).
In the best case of their experiments they achieve a median KL-divergence of approximately $0.1$.
This suggests that Algorithm~\ref{alg:cstreelearn} is quite competitive even with exact search methods that do not scale to large systems.

\begin{figure}[t]
	\centering
	\includegraphics[width=0.5\linewidth]{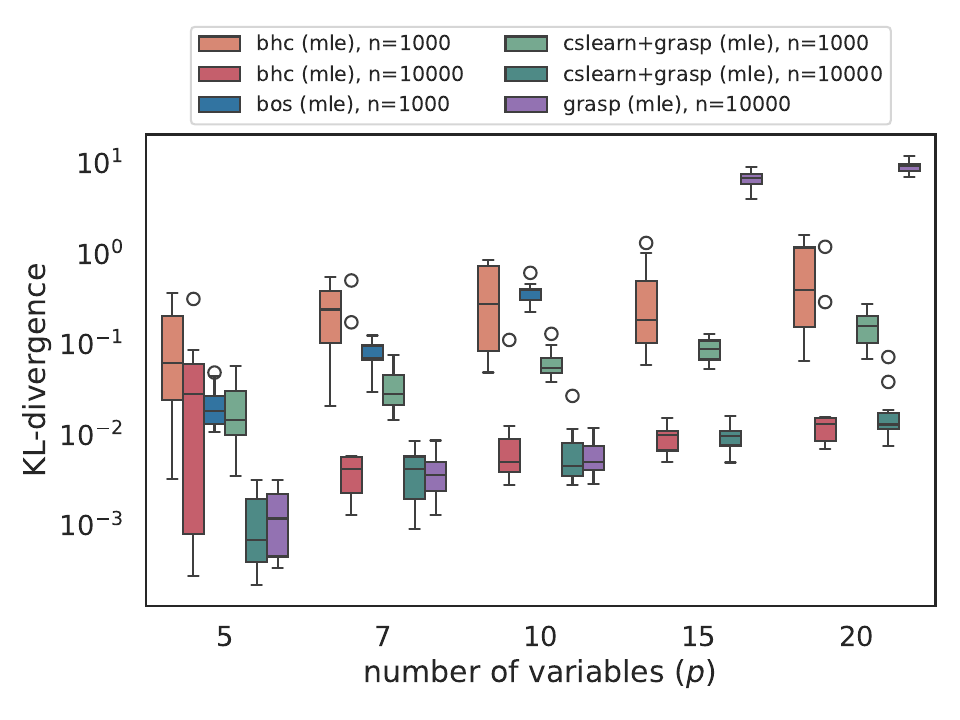}
	\caption{Accuracy comparison of CSlearn with two staged tree algorithms.
		Plots present results on a semilog scale.
	}
	\label{fig:kl-div-2c}
\end{figure}

We used the staged trees \texttt{R} package \citep{stagedtrees} to provide comparisons of CSlearn with staged tree learning algorithms.
In Figure~\ref{fig:kl-div-2c}, we compare the performance of CSlearn using GRaSP in Phase~1 with two staged tree learning algorithms, Best Order Search and Backwards Hill-Climbing, on sample sizes $n \in \{1000, 10000\}$.
We further used the maximum likelihood parameter estimator for CSlearn, as this is the standard estimator for the staged tree algorithms (which do not return posteriors).
Best Order Search failed to run in reasonable time for $p\geq 15$, so results are only included for this algorithm up to $p = 10$ for sample size $n = 1000$.

Backwards Hill Climbing is a method that first estimates a Markov equivalence class and then refines a chosen directed acyclic graph in the class to an optimal staged tree.
To provide a fair comparision, we performed Backwards Hill Climbing using GRaSP to identify the DAG.
Note that Backwards Hill Climbing uses the result of GRaSP to produce a staged tree submodel of the DAG model.
This is distinct from CSlearn, which only uses the CPDAG produced by GRaSP to estimate sets of possible context-variables.
Specifically, the final result of CSlearn need not be a submodel of the model learned by GRaSP; i.e., the graph in the LDAG learned by CSlearn may have different (non-Markov equivalent) edge structure than the graph learned in this phase.

On the smaller sample size, $n=1000$, the results in Figure~\ref{fig:kl-div-2c} show CSlearn outperforming both staged tree algorithms for all $p$.
For larger $n$ and $p$ (e.g., $n = 10000, p = 15, 20$), it appears that Backwards Hill Climbing is about equally as accurate as CSlearn, with CSlearn achieving a slightly better median value.
However, it is important to note that the model learned by the staged tree algorithms need not be a CStree, or even an LDAG.
Thus, the learned staged tree may not admit the interpretable representation of an LDAG.
Given the approximately equal accuracy performance, CSlearn may then be preferable to Backwards Hill Climbing, especially when a more compact and human-interpretable representation of the data is desired.

\subsection{Scalability Evaluation}\label{subsec:scalability}

\begin{figure}[b]
	\begin{subfigure}[b]{0.45\textwidth}
		\centering
		\includegraphics[width=\linewidth]{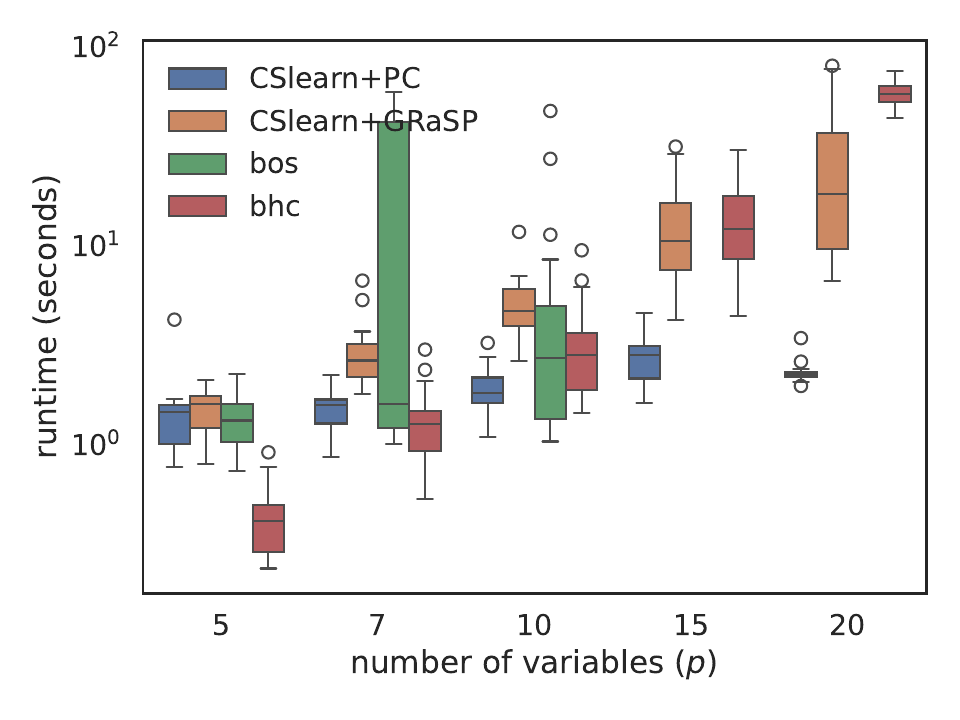}
		\caption{}
		\label{fig:runtime-3a}
	\end{subfigure}
	\hfill
	\begin{subfigure}[b]{0.45\textwidth}
		\centering
		\includegraphics[width=\linewidth]{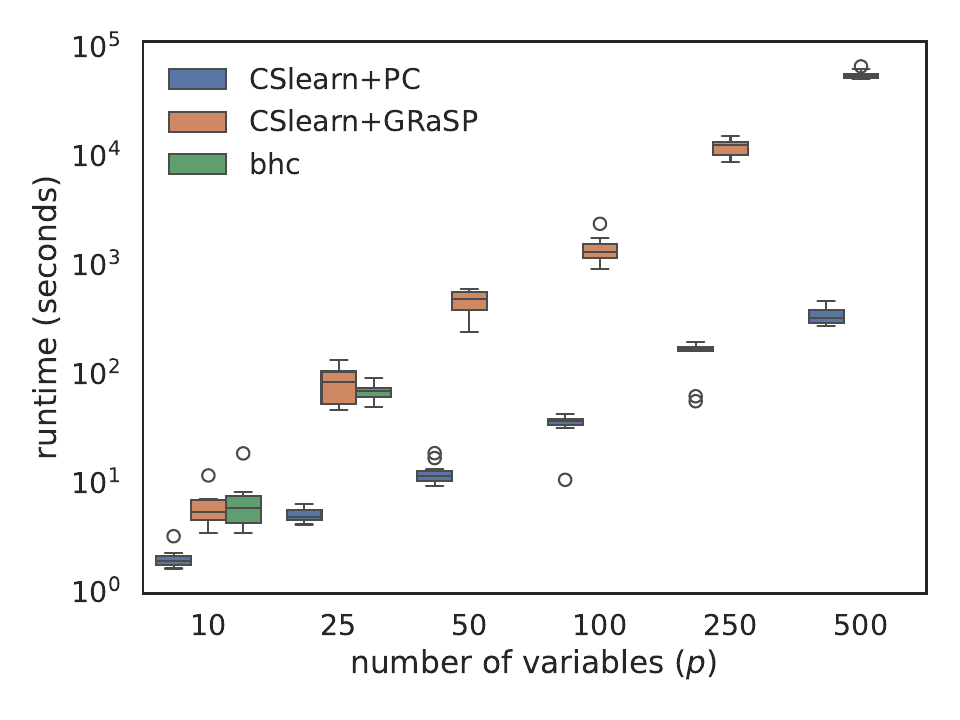}
		\caption{}
		\label{fig:runtime-3b}
	\end{subfigure}
	\caption{Runtime results.}
	\label{fig:runtime}
\end{figure}

To empirically evaluate scalability, we considered binary CStree models for $p\in\{5,7,10,15,20,50,100,250,500\}$ with $\beta = 2$.
For $n=1000$, we generated $10$ such random CStrees for each $p$, and recorded the runtime in seconds for CSlearn with PC and GRaSP used in Phase~1 and for Best Order Search and Backwards Hill Climbing.
The results are in Figure~\ref{fig:runtime}.

The results for $p \in \{5,7,10,15,20\}$ are isolated in Figure~\ref{fig:runtime-3a}.
Again, Best Order Search failed to run in reasonable time for $p \geq 15$ (e.g., more than $60$ hours without finishing), so these results are not included.
Starting at $p = 10$, we also see that the runtime for Backwards Hill Climbing begins to blow-up exponentially.
Even more, we observe that the implementation of Backwards Hill Climbing fails when increasing from \(p=25\) to \(p=50\), presumably from excessive memory usage, despite being allotted \SI{64}{\giga\byte} of RAM shared across 10 parallel processes.
On the other hand, for $p = 50$, Figure~\ref{fig:runtime-3b} shows CSlearn with PC taking approximately $10$ seconds and CSlearn with GRaSP taking about $10$ minutes.
Given the approximately equal accuracy performance of CSlearn with GRaSP and Backwards Hill Climbing for larger $p$ observed in Figure~\ref{fig:kl-div-2c}, these runtimes further purport CSlearn as a good alternative to Backwards Hill Climbing.

The runtime differences between CSlearn with PC and CSlearn with GRaSP reflects the computational efficiency of the directed acyclic graph learning algorithm used in Phase~1.
A more efficient choice of algorithm in Phase~1 naturally leads to CSlearn being more efficient.
For instance, for $p = 500$, CSlearn with GRaSP takes about $27$ hours whereas CSlearn with PC takes approximately $7$ minutes.
Given that Figure~\ref{fig:kl-div} suggests that CSlearn with GRaSP performs better than CSlearn with PC in terms of accuracy, a trade-off between runtime and accuracy may need to be made for very large systems ($p\approx 500$).
On the other hand, CSlearn with GRaSP runs in about $15$ minutes graphs with $p = 100$ variables, suggesting that no such trade-off need be made for reasonably large graphs.

Moreover, when considered in regards to the accuracy of CSlearn observed in Figure~\ref{fig:kl-div}, this runtime is quite good compared to previous methods with high accuracy, such as the exact search methods of \citet{hyttinen2018structure} where runtimes are on the order $7$ hours even for $p \approx 40$.

\subsection{Real World Example: ALARM}\label{subsec:realdata}
\begin{figure}
	\centering
	\footnotesize
	% \includesvg[width=0.9\linewidth]{source_alarm_ldag_repr_13_0}
	\resizebox{\linewidth}{!}{\huge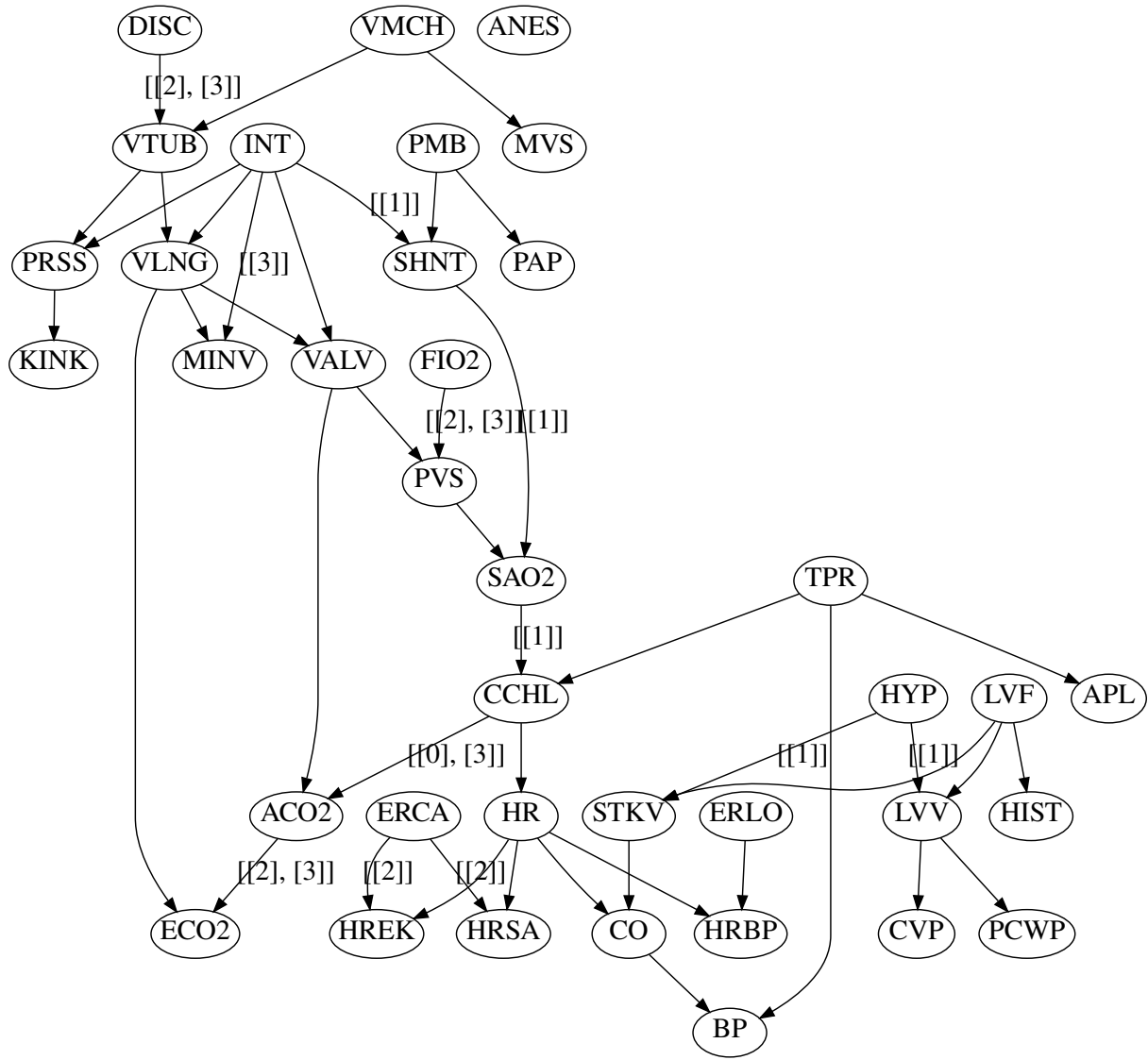}
	\caption{LDAG of the CStree learned for the ALARM data.}
	\label{fig:realdata}
\end{figure}

As a benchmark example for potential real data applications, we ran CSlearn with PC on the ALARM data set available in the \texttt{bnlearn} package in \texttt{R}.
The data set consists of $20000$ joint samples from $37$ categorical variables including binary, ternary and quartic variables (see the Supplement \citep{RMS2024supplement}).
The LDAG representation of the learned model is presented in Figure~\ref{fig:realdata}.
With $41$ edges, the underlying directed acyclic graph is only slightly sparser than the benchmark graph, which has $46$ edges.
The estimated CStree model captures $16$ CSI relations that cannot be captured by a directed acyclic graphical model of the data.
These context-specific relations are encoded in the $12$ edge labels in Figure~\ref{fig:realdata}.
The corresponding context-specific relations are presented in the Supplement \citep{RMS2024supplement}.

Algorithm~\ref{alg:cstreelearn} returned this model in $14.188$ seconds.
\citet{hyttinen2018structure} used exact optimization to estimate an LDAG representation for the same data set, which took approximately $7$ hours.
Since the benchmark DAG has nodes with parent sets bounded by only $4$, future work could aim to generalize Theorem~\ref{thm:enumeratingstagings} to $\beta = 4$, allowing us to learn denser CStree models for the data set with likely only a marginal increase in time over the current $14.188$ seconds.

%---FIGURE: ALARM---

\begin{remark}
	\label{rem:ALARM}
	We included this example since it offers a nice comparison with the results of \citet{hyttinen2018structure} for the same data set.
	Since the exact optimization methods of \citet{hyttinen2018structure} are not easily reproducible (see also Remark~\ref{rem:choice of KL}), we suggest comparing the result in Figure~\ref{fig:realdata} with \citep[Figure~4]{hyttinen2018structure}.
	Unfortunately, \citet{hyttinen2018structure} does not provide a description of how the numeric indices in their figure correspond to the variables.
	On the other hand, we may compute the accuracy of the graph in the LDAG learned by CSlearn; i.e., the proportion of off-diagonal entries of the adjacency matrix that agree with that of the ground truth.
	We find the correctly estimated edges as $TP=33$, and the incorrectly estimated edges as $FP=7$.
	Scaling by the number of edges in the true graph ($40$), we obtain $TP/40=0.825$ and $FP/40=0.175$.
	The accuracy of the (unlabeled) graph learned by CSlearn is $(TP+TN)/(37^2 - 37)=0.9895$, where $37^2- 37$ is the number of off-diagonal entries of the adjacency matrix and $TN$ is the number of correctly estimated non-edges, respectively.
\end{remark}

\subsection{Real Data Example: Mushroom}
\label{sec:real-world-example}

We demonstrate the use of CStrees for a prediction task on a real dataset \citep{misc_mushroom_73}.
The dataset consists of 22 categorical features (e.g., `cap-color' or `habitat') with described state spaces of size up to 12 as well as a binary target encoding whether each mushroom is edible or possibly poisonous.
We preprocess the data by dropping one feature (`veil-type') whose actual state space is found to be of size one.
We further removed all observations with missing values.
This leaves a sample of 5644 observations, from which we perform a 70/30 train/test split.

To learn the CStree model with Algorithm~\ref{alg:cstreelearn}, we use GRaSP \citep{lam2022greedy,zheng2024causal} to select the possible context variables in Phase~1, and we take the maximum a posteriori estimate of the stage parameters in Phase~3.
Total runtime is less than 5 minutes.
The LDAG representation of the learned CStree is shown in Figure~\ref{fig:mushroom-ldag}, with contexts corresponding to 32 edge labels a--F given in the Supplement \citep{RMS2024supplement}.
We use the learned CStree to predict the outcomes of the target variable on the held out test set, taking less than a second of runtime and achieving an accuracy of over 95.8\%.
The confusion matrix is shown in Table~\ref{tab:conf-mat-mush}, where entry \textit{row, col} reports the number (rate) of observations in class \textit{row} predicted to be in class \textit{col}.
For example, 1023 (97.2\% of) truly poisonous mushrooms are correctly predicted to be poisonous.

\begin{figure}
	\centering
	\footnotesize
	% \includesvg[width=\linewidth]{mushroom-ldag}
	\resizebox{\linewidth}{!}{\huge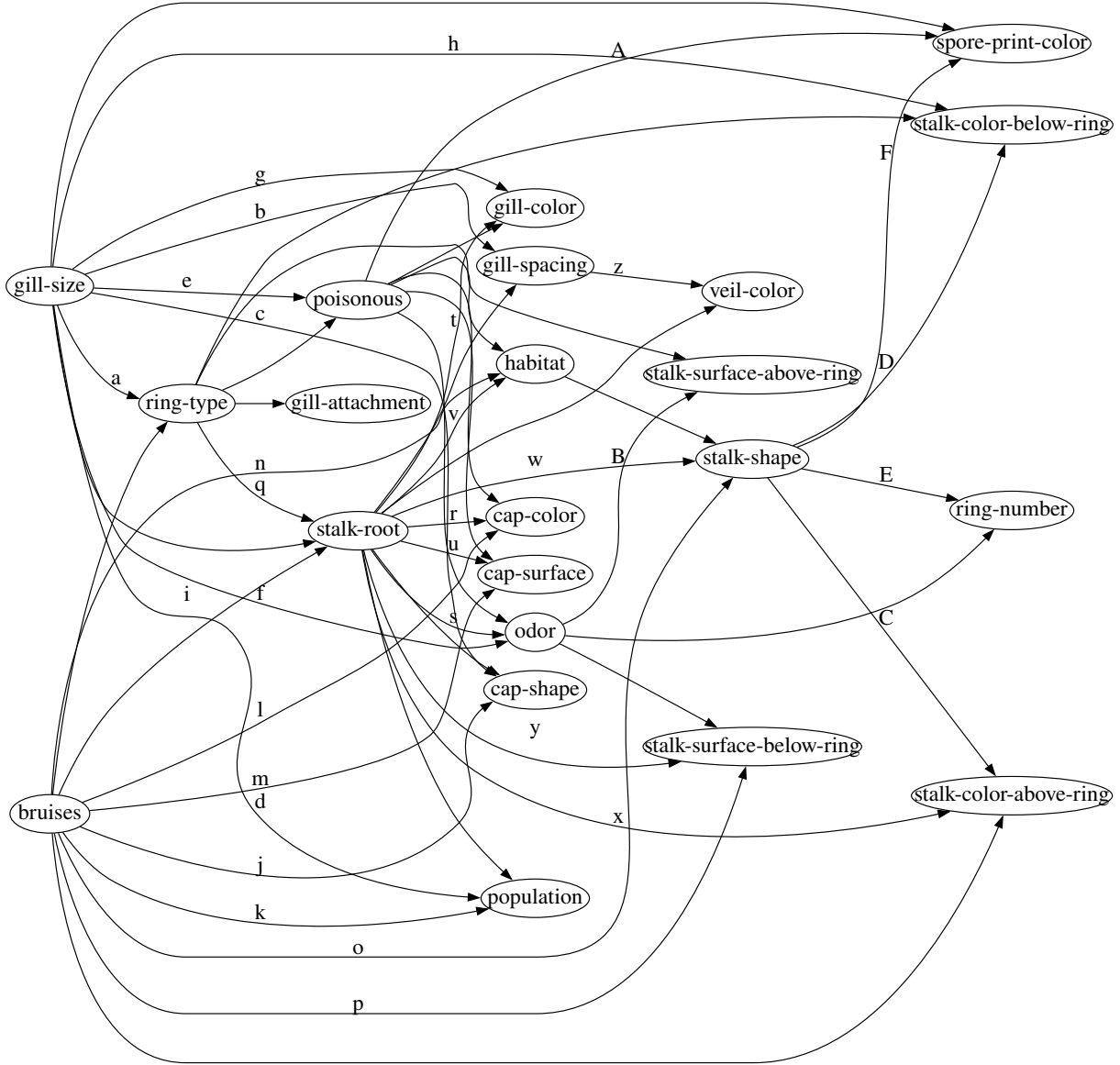}
	\caption{LDAG of the CStree learned for the Mushroom data.}
	\label{fig:mushroom-ldag}
\end{figure}

\begin{table}
	\centering
	\caption{Confusion matrix for CStree prediction on Mushroom dataset.
	}
	\label{tab:conf-mat-mush}
	\begin{tabular}{r c c}
		\toprule
		                & \multicolumn{2}{c}{predicted}                 \\ \cmidrule(r){2-3}
		                & poisonous                     & edible        \\  \midrule
		truly poisonous & 1023 (97.2\%)                 & 30 (2.8\%)    \\
		truly edible    & 41 (6.4\%)                    & 600  (93.6\%) \\\bottomrule
	\end{tabular}
\end{table}

Besides demonstrating scalable and accurate prediction with CStrees, this example illustrates how our sparsity constraint and complexity results work in practice.
In particular, notice that a given node may have a parent set size in the LDAG representation (e.g., \(|\mathrm{pa}(\text{`odor'})| = 3\)) that is larger than \(\beta=2\), which bounds the size of its corresponding stage defining contexts in the CStree---this shows that our context-specific sparsity constraint in Assumption~\ref{ass:CS} distinguishes itself from classic graph sparsity (Assumption~\ref{ass:DAG}).
Furthermore, for \(\beta = 2\), our complexity result (Theorem~\ref{thm:loscorecomp}) says that the main factors influencing time complexity are state space sizes and the maximum size of the sets of possible context variables set used (i.e., \(|K|\)).
Notice that the ALARM example (Section~\ref{subsec:realdata}), which had \(|K|=4\) and maximum state space size equaling 5, ran in about 14 seconds, while this Mushroom example, which has \(|K|=5\) and maximum state space size (after preprocessing) equaling 9, runs in around 5 minutes---this shows the importance of the \(|K|\) and maximum state space size values for runtime, as indicated in Theorem~\ref{thm:loscorecomp}.

\begin{remark}
	In the Supplement \citep{RMS2024supplement}, we also apply CSlearn with PC to the Sachs protein-signaling network data set \citep{sachs2005causal}.
	The results for the Sachs data set also demonstrate how Algorithm~\ref{alg:cstreelearn}, CStrees and their LDAG representations can efficiently discover and represent context-specific relations in the data that are obscured by a directed acyclic graph.
\end{remark}

\section{Discussion}\label{sec:future}
CSlearn (Algorithm~\ref{alg:cstreelearn}) offers an accurate and scalable method for estimating compact representations of context-specific conditional independence structure in the form of CStrees.
To achieve these features, we rely on the novel context-specific sparsity constraint $\beta$ introduced in Assumption~\ref{ass:CS}, which is specific to the definition of a CStree model via its generalization of the factorization definition~\eqref{eqn:factorization} of a directed acyclic graphical model.
While CStrees admit the desirable LDAG representations, this sparsity bound is not readily applicable to general LDAGs due to their definition via pairwise CSI relations.
It would be interesting to see if Assumption~\ref{ass:CS} extends to the more general family of LDAGs, allowing for a generalization of Algorithm~\ref{alg:cstreelearn} to this broader family of models.

Our implementation of CSlearn is only possible due to the result of Theorem~\ref{thm:enumeratingstagings}, which enumerates CStrees whose stages are defined by sets of context-variables with at most $\beta = 2$ elements.
This sparsity constraint $\beta = 2$ shows efficient and accurate performance in the experimental and real data examples in Section~\ref{sec:learningex}.
Larger values of $\beta$ would naturally allow for increasingly rich context-specific structure, but this requires generalizing Theorem~\ref{thm:enumeratingstagings} and hence solving increasingly challenging cases of the combinatorial problem of \citet{alon2023partitioning}.
A solution to their problem for arbitrary $\beta$ seems difficult.
However, based on the experimental results, perhaps a reasonable goal is to generalize Theorem~\ref{thm:enumeratingstagings} to $\beta \leq 10$, to match reasonable choices of $\alpha$.

Other future work could include extending Algorithm~\ref{alg:cstreelearn} to support learning interventional CStree models as described in \citep{duarte2021representation}, or to general stage trees where one need not solve the problem of \citet{alon2023partitioning} to achieve model enumeration.
We note, however, that in this latter regime, we would lose the compact LDAG representation of the estimated models.

\subsection*{Acknowledgements}
F.~L.~Rios and L.~Solus were partially supported by the G\"oran Gustafsson Foundation.
A.~Markham and L.~Solus were partially supported by the Wallenberg Autonomous Systems and Software Program (WASP) funded by the Knut and Alice Wallenberg Foundation.
L.~Solus was additionally supported by the G\"oran Gustafsson Prize for Young Researchers, a project grant from KTH Digital Futures and a Starting Grant from The Swedish Research Council.
We would also like to thank Danai Deligeorgaki for valuable input and fruitful discussions.

\bibliographystyle{plainnat}
\bibliography{main}

\clearpage

\thispagestyle{empty}

\appendix

\begin{center}{\large SUPPLEMENT TO: SCALABLE STRUCTURE LEARNING FOR SPARSE CONTEXT-SPECIFIC SYSTEMS}
\end{center}

% \section{Introduction}

This supplement contains additional technical details regarding the models studied in the main paper ``Scalable learning for sparse context-specific systems.
''
All non-local references refer to results in the main paper and inherit the same numbering.
It contains the proofs of \textcolor{cyan}{Theorem~1} (see Section~\ref{sec:proof}) and \textcolor{cyan}{Theorem~2} (see Section~\ref{sec:complexity}).
Section~\ref{sec:representations} contains additional technical details and examples regarding the construction of the various graphical representations of CStrees.
Section~\ref{sec:additionalexp} contains additional experimental results and supporting details for the experiments presented in the main paper.

\section{Recovering the different graphical representations of a CStree}
\label{sec:representations}

\subsection{Context-specific Conditional Independence}

Recall from \textcolor{cyan}{Section~3} that a CStree for categorical variables $(X_1,\ldots,X_p)$ is an ordered pair $\mathcal{T} = (\pi, \mathbf{s})$ where $\pi$ is a variable ordering and $\mathbf{s}$ is a collection of sets called the stages.
% Section~\ref{sec:cstrees}
The easiest representation to produce directly from the pair $(\pi, \mathbf{s})$ is the staged tree representation, whose construction is detailed in \textcolor{cyan}{Section~3}.
% Section~\ref{sec:cstrees}.
While the staged tree representation is perhaps the most direct graphical representation of the pair $(\pi,\mathbf{s})$, they are typically large graphs with numerous colored nodes that can make them difficult to directly interpret or manipulate.
More interpretable representations include the LDAG representation of \citet{pensar2015labeled} and the minimal context graph representation of \citet{duarte2021representation}.
The construction of these two representations from the pair $(\pi,\mathbf{s})$ is described below.
Both make use of the so-called \emph{context-specific conditional independence axiom} presented in \citep{corander2019logical} and \citep{duarte2021representation}:
\begin{itemize}
	\item (Symmetry) If $\vX_A \independent \vX_B | \vX_C, \vX_S = \vx_S$ then $\vX_B \independent \vX_A | \vX_C, \vX_S = \vx_S$.
	\item (Decomposition) If $\vX_A \independent \vX_{B\cup D} | \vX_C, \vX_S = \vx_S$ then $\vX_A \independent \vX_B | \vX_C, \vX_S = \vx_S$.
	\item (Weak Union) If $\vX_A \independent \vX_{B\cup D} | \vX_C, \vX_S = \vx_S$ then $\vX_A \independent \vX_{B} | \vX_{C\cup D}, \vX_S = \vx_S$.
	\item (Contraction) If $\vX_A \independent \vX_{B} | \vX_{C\cup D}, \vX_S = \vx_S$ and $\vX_A \independent \vX_{D} | \vX_{C}, \vX_S = \vx_S$ then $\vX_A \independent \vX_{B\cup D} | \vX_{C}, \vX_S = \vx_S$
	\item (Intersection) If $\vX_A \independent \vX_{B} | \vX_{C \cup D}, \vX_S = \vx_S$ and $\vX_A \independent \vX_{D} | \vX_{C\cup B}, \vX_S = \vx_S$ then $\vX_A \independent \vX_{B\cup D} | \vX_C, \vX_S = \vx_S$.
	\item (Specialization) If $\vX_A \independent \vX_B | \vX_C, \vX_S = \vx_S$ and  $D\subseteq C$ then $\vX_A \independent \vX_B | \vX_{C\setminus D}, \vX_{S\cup D} = \vx_S\vx_D$ for all $\vx_D\in\mathcal{X}_D$.
	\item (Absorption) If $\vX_A \independent \vX_B | \vX_C, \vX_S = \vx_S$ and $D\subseteq S$ such that $\vX_A \independent \vX_B | \vX_C, \vX_S = \vx_{S\setminus D}\vy_D$ for all $\vy_D\in\mathcal{X}_D$ then $\vX_A \independent \vX_B | \vX_{C\cup D}, \vX_{S\setminus D} = \vx_{S\setminus D}$.
\end{itemize}
The above implications hold for all positive probability distributions.

\subsection{The LDAG representation}
\label{subsec:LDAGrep}
We refer the reader first to the definition of an LDAG given in \textcolor{cyan}{Section~3} before proceeding.
Given a CStree $\mathcal{T} = (\pi, \mathbf{s})$ the set of stages is a disjoint union $ \mathbf{s} = \mathbf{s}_i \cup \cdots \cup \mathbf{s}_p $ where
\begin{displaymath}
	\mathbf{s}_{i} = \{\mathcal{S}_{\pi,i}(\vx_S) : X_{\pi_i} \independent \vX_{[\pi_1:\pi_{i-1}]\setminus S} | \vX_S = \vx_S\in\mathcal{C}_{\pi,i}\}.
\end{displaymath}
Each set $\mathcal{S}_{\pi,i}(\vx_S)$ consists of all joint outcomes $\vx_{[\pi_1:\pi_{i-1}]}\in\mathcal{X}_{[\pi_1:\pi_{i-1}]}$ that agree with the context $\vx_S$ when restricted to the values $x_i$ for $i\in S$.
Hence, from the elements of the set $\mathcal{S}_{\pi,i}(\vx_S)$ we can directly deduce the context $\vX_S = \vx_S$ and recover the CSI relation
\begin{displaymath}
	X_{\pi_i} \independent \vX_{[\pi_1:\pi_{i-1}] \setminus S} | \vX_S = \vx_S.
\end{displaymath}
Doing this for all $\mathcal{S}_{\pi,i}(\vx_S)\in\mathbf{s}_i$ produces the set of CSI relations
\begin{displaymath}
	\mathcal{C}_{\pi,i} = \{X_{\pi_i} \independent \vX_{[\pi_1:\pi_{i-1}] \setminus S} | \vX_S = \vx_S \mbox{ entailed by $\vX$}\}.
\end{displaymath}
Taking the union over the sets $S\subseteq [\pi_1:\pi_{i-1}]$ for which there is a relation in $\mathcal{C}_{\pi, i}$ with context $\vX_S = \vx_S$, we obtain a set $\Pi_i(\pi_i)$.
Let $\mathcal{G} = ([p], E)$ be the directed acyclic graph in which $\pa_{\mathcal{G}}(\pi_i) = \Pi_i(\pi_i)$ for all $i\in[p]$.
It follows that $\vX$ is Markov to $\mathcal{G}$, which we call a (minimal) I-MAP of the CStree $\mathcal{T}$.

\begin{example}
	\label{ex:appCStree}
	Consider the CStree defined by the collection of CSI relations
	\begin{equation*}
		\begin{split}
			\mathcal{C}_{\mathcal{T}} = \{ & X_2 \independent X_1, \\ &X_3 \independent X_1 | X_2 = 1,\\ &X_4 \independent X_2 | X_{1,3} = (1,1), \\ &X_4 \independent X_2 | X_{1,3} = (1,0),\\ &X_4 \independent X_2 | X_{1,3} = (0,1)\}.
		\end{split}
	\end{equation*}
	Its staged tree representation is
	\begin{center}
		\begin{tikzpicture}[thick,scale=0.15]
			%---NODES---
			\node[draw, fill=black!0, inner sep=2pt, rounded corners, minimum width=2pt] (w3) at (6,15)  {\scriptsize 1111};
			\node[draw, fill=black!0, inner sep=2pt, rounded corners, minimum width=2pt] (w4) at (6,13.5) {\scriptsize 1110};
			\node[draw, fill=black!0, inner sep=2pt, rounded corners, minimum width=2pt] (w5) at (6,12) {\scriptsize 1101};
			\node[draw, fill=black!0, inner sep=2pt, rounded corners, minimum width=2pt] (w6) at (6,10.5) {\scriptsize 1100};
			\node[draw, fill=black!0, inner sep=2pt, rounded corners, minimum width=2pt] (v3) at (6,9)  {\scriptsize 1011};
			\node[draw, fill=black!0, inner sep=2pt, rounded corners, minimum width=2pt] (v4) at (6,7.5) {\scriptsize 1010};
			\node[draw, fill=black!0, inner sep=2pt, rounded corners, minimum width=2pt] (v5) at (6,6) {\scriptsize 1001};
			\node[draw, fill=black!0, inner sep=2pt, rounded corners, minimum width=2pt] (v6) at (6,4.5) {\scriptsize 1000};
			\node[draw, fill=black!0, inner sep=2pt, rounded corners, minimum width=2pt] (w3i) at (6,3)  {\scriptsize 0111};
			\node[draw, fill=black!0, inner sep=2pt, rounded corners, minimum width=2pt] (w4i) at (6,1.5) {\scriptsize 0110};
			\node[draw, fill=black!0, inner sep=2pt, rounded corners, minimum width=2pt] (w5i) at (6,0) {\scriptsize 0101};
			\node[draw, fill=black!0, inner sep=2pt, rounded corners, minimum width=2pt] (w6i) at (6,-1.5) {\scriptsize 0100};
			\node[draw, fill=black!0, inner sep=2pt, rounded corners, minimum width=2pt] (v3i) at (6,-3)  {\scriptsize 0011};
			\node[draw, fill=black!0, inner sep=2pt, rounded corners, minimum width=2pt] (v4i) at (6,-4.5) {\scriptsize 0010};
			\node[draw, fill=black!0, inner sep=2pt, rounded corners, minimum width=2pt] (v5i) at (6,-6) {\scriptsize 0001};
			\node[draw, fill=black!0, inner sep=2pt, rounded corners, minimum width=2pt] (v6i) at (6,-7.5) {\scriptsize 0000};

			\node[draw, fill=blue!40, inner sep=2pt, rounded corners, minimum width=2pt] (w1) at (-2,14.25) {\scriptsize 111};
			\node[draw, fill=orange!90, inner sep=2pt, rounded corners, minimum width=2pt] (w2) at (-2,11.25) {\scriptsize 110};
			\node[draw, fill=blue!40, inner sep=2pt, rounded corners, minimum width=2pt] (v1) at (-2,8.25) {\scriptsize 101};
			\node[draw, fill=orange!90, inner sep=2pt, rounded corners, minimum width=2pt] (v2) at (-2,5.25) {\scriptsize 100};
			\node[draw, fill=red!90, inner sep=2pt, rounded corners, minimum width=2pt] (w1i) at (-2,2.25) {\scriptsize 011};
			\node[draw, fill=orange!0, inner sep=2pt, rounded corners, minimum width=2pt] (w2i) at (-2,-0.75) {\scriptsize 010};
			\node[draw, fill=red!90, inner sep=2pt, rounded corners, minimum width=2pt] (v1i) at (-2,-3.75) {\scriptsize 001};
			\node[draw, fill=violet!0, inner sep=2pt, rounded corners, minimum width=2pt] (v2i) at (-2,-6.75) {\scriptsize 000};

			\node[draw, fill=green!90, inner sep=2pt, rounded corners, minimum width=2pt] (w) at (-8,12.75) {\scriptsize 11};
			\node[draw, fill=cyan!0, inner sep=2pt, rounded corners, minimum width=2pt] (v) at (-8,6.75) {\scriptsize 10};
			\node[draw, fill=green!90, inner sep=2pt, rounded corners, minimum width=2pt] (wi) at (-8,0.75) {\scriptsize 01};
			\node[draw, fill=cyan!0, inner sep=2pt, rounded corners, minimum width=2pt] (vi) at (-8,-5.25) {\scriptsize 00};

			\node[draw, fill=yellow!60, inner sep=2pt, rounded corners, minimum width=2pt] (r) at (-14,9.75) {\scriptsize 1};
			\node[draw, fill=yellow!60, inner sep=2pt, rounded corners, minimum width=2pt] (ri) at (-14,-1.75) {\scriptsize 0};

			\node[draw, fill=black!0, inner sep=2pt, rounded corners, minimum width=2pt] (I) at (-20,3) {\scriptsize r};

			%---EDGES---
			\draw[->]   (I) --    (r) ;
			\draw[->]   (I) --   (ri) ;

			\draw[->]   (r) --   (w) ;
			\draw[->]   (r) --   (v) ;

			\draw[->]   (w) --  (w1) ;
			\draw[->]   (w) --  (w2) ;

			\draw[->]   (w1) --   (w3) ;
			\draw[->]   (w1) --   (w4) ;
			\draw[->]   (w2) --  (w5) ;
			\draw[->]   (w2) --  (w6) ;

			\draw[->]   (v) --  (v1) ;
			\draw[->]   (v) --  (v2) ;

			\draw[->]   (v1) --  (v3) ;
			\draw[->]   (v1) --  (v4) ;
			\draw[->]   (v2) --  (v5) ;
			\draw[->]   (v2) --  (v6) ;

			\draw[->]   (ri) --   (wi) ;
			\draw[->]   (ri) -- (vi) ;

			\draw[->]   (wi) --  (w1i) ;
			\draw[->]   (wi) --  (w2i) ;

			\draw[->]   (w1i) --  (w3i) ;
			\draw[->]   (w1i) -- (w4i) ;
			\draw[->]   (w2i) --  (w5i) ;
			\draw[->]   (w2i) --  (w6i) ;

			\draw[->]   (vi) --  (v1i) ;
			\draw[->]   (vi) --  (v2i) ;

			\draw[->]   (v1i) --  (v3i) ;
			\draw[->]   (v1i) -- (v4i) ;
			\draw[->]   (v2i) -- (v5i) ;
			\draw[->]   (v2i) --  (v6i) ;

			%---LABELS---
			\node at (-17.5,-9) {$X_1$} ;
			\node at (-11.5,-9) {$X_2$} ;
			\node at (-5,-9) {$X_3$} ;
			\node at (2,-9) {$X_4$} ;

		\end{tikzpicture}
	\end{center}
	To recover the minimal I-MAP $\G$ for $\mathcal{T}$, we start with the node $4$.
	There are exactly three CSI relations for the variable $X_4$:
	\begin{displaymath}
		X_4 \independent X_2 | X_{1,3} = (1,1), \qquad X_4 \independent X_2 | X_{1,3} = (1,0) \qquad \mbox{and} \qquad X_4 \independent X_2 | X_{1,3} = (0,1),
	\end{displaymath}
	and each of these relations has contexts given by the variables $\{X_1,X_3\}$.
	Hence, $\Pi(4) \supseteq \{1,3\}$.
	However, there exist white nodes in the fourth level of the tree $\mathcal{T}$; namely, $(0,0,0),(0,1,0)$.
	Hence, $\Pi(4) = \{1,2,3\}$, and the parents of $4$ in $\G$ are $\pa_{\G}(4) = \{1,2,3\}$.
	Similarly, the parents of $3$ in $\G$ are $\pa_{\G}(3) = \{1,2\}$.

	Finally, note that the relation $X_2 \independent X_1$ is a genuine CI relation, and therefore it is defined by the context $\vX_\emptyset = \vx_\emptyset$.
	Hence, the parent set of node $2$ in $\G$ is $\pa_{\G}(2) = \emptyset$.
	Since there are no relations for node $1$, we further have that the parents of node $1$ in $\G$ are $\pa_{\G}(1) = \emptyset$.
	It follows that the minimal I-MAP of the CStree $\mathcal{T}$ is
	\begin{center}
		\begin{tikzpicture}[thick,scale=0.2]

			\node[circle, draw, fill=black!
				0, inner sep=1pt, minimum width=1pt] (H1) at (3.25,8) {$1$};
			\node[circle, draw, fill=black!0, inner sep=1pt, minimum width=1pt] (B1) at (-2.25,4) {$2$};
			\node[circle, draw, fill=black!0, inner sep=1pt, minimum width=1pt] (G1) at (8.25,4) {$3$};
			\node[circle, draw, fill=black!0, inner sep=1pt, minimum width=1pt] (B2) at (3.25,0) {$4$};

			%---EDGES---
			\draw[->]   (H1) -- (G1) ;
			\draw[->]   (H1) -- (B2) ;
			\draw[->]   (B1) -- (B2) ;
			\draw[->]   (G1) -- (B2) ;
			\draw[->]   (B1) -- (G1) ;
		\end{tikzpicture}
		.
	\end{center}
\end{example}

The resulting DAG $\mathcal{G}$ is the DAG forming the basis for the LDAG representation $(\mathcal{G},\mathcal{L})$ of $\mathcal{T}$.
It remains to identify the edge labels $\mathcal{L}$.
In an LDAG, the edges $j\rightarrow i\in E$ are labeled with a set of contexts $\vx_{\pa_{\G}(i)\setminus\{j\}}$ such that $X_i \independent X_j \mid \vX_{\pa_{\G}(i)\setminus\{j\}} = \vx_{\pa_{\G}(i)\setminus\{j\}}$.
For the minimal I-MAP $\G$ of $\mathcal{T}$, we know that for each $X_{\pi_i} \independent \vX_{[\pi_1:\pi_{i-1}] \setminus S} \mid \vX_S = \vx_S\in \mathcal{C}_{\pi,i}$ the set $S$ is a subset of the set $\pa_{\G}(\pi_i)$.
Applying the weak union property for CSI relations, we know that the distributions in the CStree model $\mathcal{M}(\mathcal{T})$ satisfy the relations
\begin{equation*}
	X_{\pi_i} \independent \vX_{[\pi_1:\pi_{i-1}]\setminus (\pa_{\G}(\pi_i)\setminus\pi_j)} \mid \vX_{\pa_{\G}(\pi_i)\setminus (S\cup \pi_j)}, \vX_S = \vx_S.
\end{equation*}
Applying the decomposition property it follows that the distributions in $\mathcal{M}(\mathcal{T})$ satisfy
\begin{displaymath}
	X_{\pi_i} \independent X_{\pi_j} | \vX_{\pa_{\G}(\pi_i)\setminus (S\cup \pi_j)}, \vX_S = \vx_S
\end{displaymath}
for all $\pi_j\in [\pi_1:\pi_{i-1}]\setminus \pa_{\G}(\pi_i)$, for all contexts $\vX_S = \vx_S$ defining the CSI relations in $\mathcal{C}_{\pi,i}$.
Hence, the label for the edge $\pi_j\rightarrow \pi_i$ in $\G$ would be the set
\begin{displaymath}
	L_{\pi_j,\pi_i} = \{\vx_{\pa_{\G}(\pi_i)\setminus (S\cup\pi_j)}\vx_S : \vX_S = \vx_S \mbox{ defines a relation in $\mathcal{C}_{\pi,i}$ and } \vx_{\pa_{\G}(\pi_i)\setminus (S\cup\pi_j)} \in \mathcal{X}_{\pa_{\G}(\pi_i)\setminus (S\cup\pi_j)}\}.
\end{displaymath}
The set $L_{\pi_j,\pi_i}$ can be very large.
To simplify its representation it is typical to use a $\ast$ symbol in the coordinate for outcome $x_k$ to indicate that the context is in the set for all $x_k\in\mathcal{X}_k$.

\begin{example}[Example~\ref{ex:appCStree} continued.]
	% \paragraph{Example 1 continued.}
	We first compute the edge label $L_{2,4}$ for the DAG $\G$ in Example~1.
	Since we know that the CSI relations
	\begin{displaymath}
		X_4 \independent X_2 \mid X_{1,3} = (1,1), \qquad X_4 \independent X_2 \mid X_{1,3} = (1,0) \qquad \mbox{and} \qquad X_4 \independent X_2 \mid X_{1,3} = (0,1),
	\end{displaymath}
	hold, we obtain the edge label
	\begin{displaymath}
		L_{2,4} = \{(0,1),(1,0),(1,1)\}\subset \mathcal{X}_{\pa_{\G}(4)\setminus\{2\}}.
	\end{displaymath}
	Since $(0,1)$ and $(1,1)$ are both in $L_{2,4}$ we can represent this pair of outcomes more simply as $(\ast,1)$.
	Similarly, the pair $(1,0)$ and $(1,1)$ can be represented as $(1,\ast)$.
	(Note that this representation redundantly represents the single outcome $(1,1)$, but it results in a simpler edge label:
	\begin{displaymath}
		L_{2,4} = \{(1,\ast),(\ast,1)\}.
	\end{displaymath}
	This is the edge label of the edge $2 \rightarrow 4$ in the LDAG in below.
	The remaining edge labels can be calculated analogously, yielding the LDAG representation of the CStree $\mathcal{T}$ in Example~\ref{ex:appCStree}:
	\begin{center}
		\begin{tikzpicture}[thick,scale=0.2]

			\node[circle, draw, fill=black!
				0, inner sep=1pt, minimum width=1pt] (H1) at (3.25,8) {$1$};
			\node[circle, draw, fill=black!0, inner sep=1pt, minimum width=1pt] (B1) at (-2.25,4) {$2$};
			\node[circle, draw, fill=black!0, inner sep=1pt, minimum width=1pt] (G1) at (8.25,4) {$3$};
			\node[circle, draw, fill=black!0, inner sep=1pt, minimum width=1pt] (B2) at (3.25,0) {$4$};

			%---EDGES---
			\draw[->]   (H1) -- node[midway,sloped,above]{\tiny${\{1\}}$}(G1) ;
			\draw[->]   (H1) -- (B2) ;
			\draw[->]   (B1) -- node[align=center,below, rotate=-40]{\tiny{$\{(1,\ast),$} \\ \tiny{$(\ast,1)\}$}} (B2) ;
			\draw[->]   (G1) -- (B2) ;
			\draw[->]   (B1) -- (G1) ;
		\end{tikzpicture}
		.
	\end{center}
\end{example}

Note that if an edge label $L_{j,i}$ is the emptyset, we simply do not draw it.
This indicates that the dependency represented by this edge vansishes under no contexts.

\subsection{The Minimal Context Graph Representation}

An alternative to the LDAG representation of a CStree is the representation of the model via a collection of DAGs, one for each element of a set of \emph{minimal contexts}.
A \emph{minimal context} is defined in the following way: Let $\mathcal{J}(\mathcal{T})$ denote the complete set of CSI relations that are implied by those defining the model $\mathcal{M}(\mathcal{T})$ via repeated application of the context-specific conditional independence axioms above.
Let $\vX_A \independent \vX_B | \vX_C, \vX_D = \vx_D$ be any relation in $\mathcal{J}(\mathcal{T})$.
We may apply the absorption axiom to this relation until the cardinality of the variables $D$ defining the context can no longer be reduced.
The contexts obtained by doing this for all possible elements of $\mathcal{J}(\mathcal{T})$ in all possible ways of applying absorption is necessarily a finite set of contexts, called the minimal contexts.
For each such context, $\vX_D = \vx_D$, we can consider the I-MAP of the CSI relations with this context that live in $\mathcal{J}(\mathcal{T})$.
This is a DAG whose d-separations encode the CSI relations with context $\vX_D = \vx_D$.
For example, the minimal context graph representation of the CStree depicted in Example~\ref{ex:appCStree} is
\begin{center}
	\begin{tikzpicture}[thick,scale=0.22] \draw (-10.5,3) -- (26,3) -- (26, -6) -- (-10.5, -6) -- (-10.5,3) -- cycle; \draw (-2,3) -- (-2,-6) ; \draw (7.5,3) -- (7.5,-6) ; \draw (17,3) -- (17,-6) ;

		\node[circle, draw, fill=black!
			%---NODES---
			0, inner sep=1pt, minimum width=1pt] (E1) at (0.25 - 9.5,0) {$1$};
		\node[circle, draw, fill=black!0, inner sep=1pt, minimum width=1pt] (E2) at (3.25- 9.5 + 3,0) {$2$};
		\node[circle, draw, fill=black!0, inner sep=1pt, minimum width=1pt] (E3) at (0.25- 9.5,-4) {$3$};
		\node[circle, draw, fill=black!0, inner sep=1pt, minimum width=1pt] (E4) at (6.25- 9.5,-4) {$4$};

		\node[circle, draw, fill=black!0, inner sep=1pt, minimum width=1pt] (H1) at (3.25,0) {$2$};
		\node[circle, draw, fill=black!0, inner sep=1pt, minimum width=1pt] (B1) at (0.25,-4) {$3$};
		\node[circle, draw, fill=black!0, inner sep=1pt, minimum width=1pt] (G1) at (6.25,-4) {$4$};

		\node[circle, draw, fill=black!0, inner sep=1pt, minimum width=1pt] (H2) at (12.5,0) {$1$};
		\node[circle, draw, fill=black!0, inner sep=1pt, minimum width=1pt] (B2) at (9.5,-4) {$3$};
		\node[circle, draw, fill=black!0, inner sep=1pt, minimum width=1pt] (G2) at (15.5,-4) {$4$};

		\node[circle, draw, fill=black!0, inner sep=1pt, minimum width=1pt] (H3) at (21.5,0) {$1$};
		\node[circle, draw, fill=black!0, inner sep=1pt, minimum width=1pt] (B3) at (18.5,-4) {$2$};
		\node[circle, draw, fill=black!0, inner sep=1pt, minimum width=1pt] (G3) at (24.5,-4) {$4$};

		%---EDGES---
		\draw[->]   (E1) -- (E3) ;
		\draw[->]   (E1) -- (E4) ;
		\draw[->]   (E2) -- (E3) ;
		\draw[->]   (E2) -- (E4) ;
		\draw[->]   (E3) -- (E4) ;

		\draw[->]   (H1) -- (B1) ;
		\draw[->]   (B1) -- (G1) ;

		\draw[->]   (H2) -- (G2) ;
		\draw[->]   (B2) -- (G2) ;

		\draw[->]   (H3) -- (B3) ;
		\draw[->]   (H3) -- (G3) ;

		%---LABELS---
		\node at (-7.5,2) {$\G_{X_\emptyset = x_\emptyset}$} ;
		\node at (0.5,2) {$\G_{X_1 = 1}$} ;
		\node at (10,2) {$\G_{X_2 = 1}$} ;
		\node at (19.5,2) {$\G_{X_3 = 1}$} ;
	\end{tikzpicture}
\end{center}
It is shown in \citep{duarte2021representation} that two CStrees define the same set of distributions (i.e. are Markov equivalent) if and only if they have the same set of minimal contexts and each context gives a pair of Markov equivalent DAGs.
This makes this representation useful for deducing model equivalence.
However, it is currently an intractable representation to compute for large systems of variables.
Hence, we use the LDAG representation throughout this paper.

\section{Enumerating the Possible Stagings for Sparse CStrees}
\label{sec:proof}

In this section, we describe a method for producing a list of all possible CStrees $\mathcal{T} = (\pi,\mathbf{s})$ with stages defined by contexts $\vx_S$ with $|S| \leq 2$ for distributions $(X_1,\ldots,X_p)$ with state space $\mathcal{X} = \prod_{i=1}^p[d_i]$ and $d_1,\ldots, d_p >1$.
(That is, all CStrees satisfying \textcolor{cyan}{Assumption~1} with $\beta = 2$.)
This is a necessary computation for computing the local and order scores
in~\textcolor{cyan}{(8)} and~\textcolor{cyan}{(6)}
that are used in both the stochastic search and exact optimization phases of the algorithm.
Since the variable ordering is fixed, we assume in the following that it is the natural ordering $\pi = 12\cdots p$ and build a list of possible stagings for each level $\mathcal{X}_{[i]}$.
Each staging corresponds to a way to color the nodes $\mathcal{X}_{[i]}$ in the staged tree representation of the CStree given in \textcolor{cyan}{Figure~1a}.
% Figure~\ref{fig:cstree}.
In the following, we refer to the set of nodes $\mathcal{X}_{[i]}$ as \emph{level $i$} of the CStree.
Identifying all colorings of level $i$ that yield a CStree in turn yields a way to construct all CStrees since any one stage only contains vertices of the tree in a single level.

Recall that a stage $\mathcal{S}(\vx_S)\subset \mathcal{X}_{[i]}$ is a subset of nodes satisfying \[ \mathcal{S}(\vx_S) = \{\vy_{[i]}\in \mathcal{X}_{[i]} : \vy_{[i]\cap S} = \vx_S\} \] for some $\vx_S\in\mathcal{X}_{S}$.
We call $\vx_S$ the \emph{stage-defining context} and $S$ the \emph{context variables} of the stage.
Recall further that a \emph{(CStree) staging} of level $i$ is a collection of disjoint stages $\mathcal{S}(\vx_S)\subset \mathcal{X}_{[i]}$.
In the definition of a stage given in \textcolor{cyan}{Section~3}, the stage $\mathcal{S}(\vx_S)$ is defined by a CSI relation $X_{i} \independent \vX_{[i-1]\setminus S} | \vX_S = \vx_S$.
% Section~\ref{sec:cstrees},
If we allow the set $[i-1]\setminus S$ to be empty in this relation (equivalently allowing $S = [i-1]$ then such a statement indicates that $X_i$ depends on all preceding variables in the order.
The corresponding stage for such a statement is the singleton stage $\{\vx_S\}$.
Considering these stages in our definition of the a staging of level $i$, we see that a staging $\mathbf{s}_i$ is a partition of the level $\mathcal{X}_{[i]}$, where the singletons in this partition are defined by contexts having a set of context variables $S$ satisfying $|S| = i-1$.
In this paper, we consider sparse CStree models where $|S|\leq 2$ for all stages in all levels.
So we would like to enumerate the stagings of level $i$ in which all sets of context variables have cardinality at most two (i.e., $|S| = 0,1,$ or $2$).

To enumerate the possible stagings of level $i$ with stages defined by only $0,1$, or $2$ context variables, we can build a list of the stagings in four separate steps:
\begin{enumerate}
	\item a list containing all stagings $\mathbf{s}_i$ of level $i$ that contain a stage with context variables $S$ satisfying $|S| = 0$.
	\item a list containing all stagings of level $i$ that contain only stages having context variables $S$ satisfying $|S| = 1$.
	\item a list containing all stagings of level $i$ that contain only stages having context variables $S$ satisfying $|S| = 2$.
	\item a list containing all stagings of level $i$ that contain at least one stage having context variables $S$ satisfying $|S| = 1$ and at least one stage having context variables $S$ satisfying $|S| = 2$.
\end{enumerate}

The following lemma addresses $(1)$:

\begin{lemma}
	\label{lem:0cvars}
	A list of the CStree stagings of level $i$ in which all stages satisfy $|S| = 0$ contains exactly one element; namely, $\mathcal{X}_{[i]}$.
\end{lemma}

\begin{proof}
	Suppose $\mathbf{s}_{i}$ is a staging of level $i$ that contains a stage $\mathcal{S}(\vx_{S})$ for which $|S| = 0$.
	It follows that $S =\emptyset$ and $\vx_S$ is the empty context.
	We then have that \[ \mathcal{S}(\vx_S) = \{\vy_{[i]}\in \mathcal{X}_{[i]} : \vy_{[i]\cap S} = \vx_S\} = \mathcal{X}_{[i]}, \] which implies that all outcomes in level $i$ are contained in a single stage; namely, $\mathcal{S}(\vx_S)$.
	Since stages in a CStree staging of level $i$ must be disjoint, it follows that this is the only such staging of level $i$.
\end{proof}

The following lemma addresses $(2)$:

\begin{lemma}
	\label{lem:1cvar}
	All CStree stagings of level $i$ in which all stages $\mathcal{S}(\vx_S)$ have $|S| = 1$ are of the form \[ \mathbf{s}_i = \{\mathcal{S}(x_j) : x_j \in\mathcal{X}_j\} \] for some $j\in[i]$.
	There are exactly $i$ stagings of level $i$ of this type.
\end{lemma}

\begin{proof}
	Let $\mathbf{s}_i$ be a CStree staging of level $i$ in which all stages $\mathcal{S}(\vx_S)\in\mathbf{s}_i$ satisfy $|S| = 1$.
	It follows that there exists at least one stage $\mathcal{S}(x_j)\in\mathbf{s}_i$ for some $j\in[i]$ and some outcome $x_j\in\mathcal{X}_j$.
	Suppose for the sake of contradiction that $\mathbf{s}_i$ contains a second stage $\mathcal{S}(x_k)$ for some $k\neq j$.
	Since \[ \mathcal{S}(x_k) = \{\vy_{[i]}\in \mathcal{X}_{[i]} : \vy_{[i]\cap \{k\}} = x_k\}, \] it follows that $\mathcal{S}(x_k)$ contains all outcomes of $X_1,\ldots, X_i$ satisfying both $X_j = x_j$ and $X_k = x_k$.
	Thus, $\mathcal{S}(x_j) \cap \mathcal{S}(x_k) \neq \emptyset$, contradicting the assumption that $\mathbf{s}_i$ is a partition of $\mathcal{X}_{[i]}$.
	Therefore, all stages $\mathcal{S}(\vx_S) \in\mathbf{s}_i$ satisfy $S = \{j\}$.

	For a fixed $j$ there is exactly one such partition of $\mathcal{X}_{[i]}$ of this type, and it is \[ \mathbf{s}_i = \{\mathcal{S}(x_j) : x_j \in\mathcal{X}_j\}.
	\]
	Moreover, such a staging for any choice of $j\in[i]$ is a valid CStree staging, completing the proof.
\end{proof}

To address $(2)$ in the above list, we require a few lemmas.

\begin{lemma}
	\label{lem:2cvars_either}
	Let $\mathbf{s}_i$ be a CStree staging of level $i$ in which all stages $\mathcal{S}(\vx_S)$ satisfy $|S| = 2$, and suppose that $\mathcal{S}(x_jx_k)\in \mathbf{s}_i$ for some $j,k\in[i]$, $x_j\in\mathcal{X}_j$ and $x_k\in\mathcal{X}_k$.
	Then any other stage $\mathcal{S}(\vx_S)\in\mathbf{s}_i$ satisfies either
	\begin{enumerate}
		\item[(i)] $k\in S$, or
		\item[(ii)] $j\in S$.
	\end{enumerate}
\end{lemma}

\begin{proof}
	Suppose, for the sake of contradiction, that there exists $\mathcal{S}(\vx_S)\in\mathbf{s}_i$ such that $S = \{t,m\}$ but $\{j,k\}\cap \{t,m\} = \emptyset$.
	Since \[ \mathcal{S}(x_tx_m) = \{\vy_{[i]}\in \mathcal{X}_{[i]} : \vy_{[i]\cap \{t,m\}} = x_tx_m\}, \] it follows that $\mathcal{S}(x_tx_m)$ contains the outcomes of $X_1,\ldots, X_i$ satisfying $X_t = x_t, X_m = x_m, X_j = x_j$ and $X_k = x_k$.
	Thus, $\mathcal{S}(x_jx_k) \cap\mathcal{S}(x_tx_m) \neq \emptyset$.
\end{proof}

\begin{lemma}
	\label{lem:2cvars_commonk}
	Let $\mathbf{s}_i$ be a CStree staging of level $i$ in which all stages $\mathcal{S}(\vx_S)$ satisfy $|S| = 2$.
	Then there exists $k\in[i]$ such that $k\in S$ for all $\mathcal{S}(\vx_S)\in\mathbf{s}_i$.
\end{lemma}

\begin{proof}
	Since $\mathcal{X}_{[i]}$ is nonempty, there exists at least one stage $\mathbf{S}(x_jx_k)\in\mathbf{s}_i$.
	By Lemma~\ref{lem:2cvars_either}, every other stage $\mathcal{S}(\vx_S)\in\mathbf{s}_i$ satisfies either $j\in S$ or $k\in S$.
	To prove the lemma, it suffices to show there cannot be stages $\mathcal{S}(x_j^\prime x_t)$ and $\mathcal{S}(x_k^\prime x_m)$ both in $\mathbf{s}_i$ where $t,m\notin \{j,k\}$.
	Suppose for the sake of contradiction these two additional stages are also in $\mathbf{s}_i$.

	Consider first that stage $\mathcal{S}(x_k^\prime x_m)$.
	We must then have that $m \neq j$, from which it follows that $x_k^\prime \neq x_k$.
	Otherwise, we would have that $\mathcal{S}(x_jx_k)\cap \mathcal{S}(x_k^\prime x_t) \neq \emptyset$ as $x_j$ varies in the outcomes contained in $\mathcal{S}(x_k^\prime x_t)$.
	It follows that all stages $\mathcal{S}(\vx_S)\in\mathbf{s}_i$ satisfying $k\in S$ have distinct $\vx_{S\cap \{k\}}\in\mathcal{X}_k$.

	By symmetry of this argument, the same is true for all stages having $j\in S$.
	Consider now the stage $\mathcal{S}(x_j^\prime x_t)$.
	It follows that $t\neq k$ and $x_j^\prime \neq x_j$.
	We claim that $t = m$.
	To see this, suppose otherwise.
	It would then follow that $\mathcal{S}(x_j^\prime x_t) \cap \mathcal{S}(x_k^\prime x_m)\neq \emptyset$ since $x_j$ and $x_t$ vary in $\mathcal{S}(x_k^\prime x_m)$, meaning that $\mathcal{S}(x_k^\prime x_m)$ contains an outcome satisfying $X_t = x_t, X_j = x_j^\prime, X_k = x_k^\prime$ and $X_m =x_m$.
	Thus, we must have that $t = m$.

	It follows that we can let $x_m = x_t^\prime$.
	We claim that we must have $x_t \neq x_t^\prime$.
	To see this, note that if $x_t = x_t^\prime$ then we would have $\mathcal{S}(x_j^\prime x_t) \cap \mathcal{S}(x_k^\prime x_t^\prime)\neq \emptyset$ since $x_j$ varies in $\mathcal{S}(x_k^\prime x_t^\prime)$.
	Hence, $\mathcal{S}(x_k^\prime x_t^\prime)$ would contain outcomes satisfying $X_k = x_k^\prime$, $X_j = x_j^\prime$ and $X_t = x_t = x_t^\prime$.

	Hence, we have three stages in $\mathbf{s}_i$: $\mathcal{S}(x_jx_k), \mathcal{S}(x_j^\prime x_t)$ and $\mathcal{S}(x_k^\prime x_t^\prime)$ where $x_j \neq x_j^\prime$, $x_k \neq x_k^\prime$ and $x_t \neq x_t^\prime$.
	Note that the outcomes of $X_1,\ldots, X_i$ satisfying $X_j = x_j^\prime, X_k = x_k$ and $X_t = x_t^\prime$ must also be contained in some stage in $\mathbf{s}_i$, but they clearly cannot be in any of these three.
	Since all stages $\mathcal{S}(\vx_S)$ in $\mathbf{s}_i$ satisfy $|S| = 2$ it follows that a stage $\mathcal{S}(\vx_S)$ containing such an outcome must satisfy one of the following three conditions
	\begin{enumerate}
		\item[(a)] $\vx_S$ contains two of the outcomes $x_j^\prime, x_k, x_t^\prime$,
		\item[(b)] $\vx_S$ contains exactly one of the outcomes $x_j^\prime, x_k, x_t^\prime$, or
		\item[(c)] $\vx_S$ contains none of $x_j^\prime, x_k, x_t^\prime$.
	\end{enumerate}
	In (c), since we are assuming the stage $\mathcal{S}(\vx_S)$ contains an outcome of $X_1,\ldots, X_i$ satisfying $X_j = x_j^\prime, X_k = x_k$ and $X_t = x_t^\prime$, it must be that $S\cap\{j,k,t\} = \emptyset$.
	However, such a stage will clearly have nonempty intersection with all three of $\mathcal{S}(x_jx_k), \mathcal{S}(x_j^\prime x_t)$ and $\mathcal{S}(x_k^\prime x_t^\prime)$, a contradiction.

	In (b), suppose without loss of generality that $x_j^\prime$ is contained in $\vx_S$.
	Since $\vx_S$ contains exactly one of $x_j^\prime, x_k, x_t^\prime$, and $\mathcal{S}(\vx_S)$ contains an outcome of $X_1,\ldots, X_i$ satisfying $X_j = x_j^\prime, X_k = x_k$ and $X_t = x_t^\prime$, it follows that $k,t\notin S$.
	It follows that $\mathcal{S}(\vx_S)\cap \mathcal{S}(x_j^\prime x_t)\neq \emptyset$, a contradiction.

	In (c), suppose, without loss of generality that $\vx_S = x_j^\prime x_k$.
	It follows that $x_t$ varies in $\mathcal{S}(\vx_S)$ and hence $\mathcal{S}(\vx_S)$ contains outcomes satisfying $X_t = x_t$, $X_j = x_j^\prime$ and $X_k = x_k$.
	Thus, $\mathcal{S}(\vx_S) \cap\mathcal{S}(x_j^\prime x_t)\neq \emptyset$, another contradiction.

	We started by assuming that $\mathbf{s}_i$ contains a stage $\mathcal{S}(x_jx_k)$ as well as stages $\mathcal{S}(x_j^\prime x_t)$ and $\mathcal{S}(x_k^\prime x_m)$ both in $\mathbf{s}_i$ where $t,m\notin \{j,k\}$.
	From this assumption we have derived the observation that these three stages are the stages $\mathcal{S}(x_jx_k), \mathcal{S}(x_j^\prime x_t)$ and $\mathcal{S}(x_k^\prime x_t^\prime)$ where $x_j \neq x_j^\prime$, $x_k \neq x_k^\prime$ and $x_t \neq x_t^\prime$, non of which contain the outcomes in $\mathcal{X}_{[i]}$ satisfying $X_j = x_j^\prime, X_k = x_k$ and $X_t = x_t^\prime$.
	However, the three contradictions above show that no stage could possible exist in $\mathbf{s}_i$ that contains these outcomes.
	Hence, we see that when $\mathbf{s}(x_jx_k)\in\mathbf{s}_i$ then there cannot also be stages $\mathcal{S}(x_j^\prime x_t)$ and $\mathcal{S}(x_k^\prime x_m)$ both in $\mathbf{s}_i$ where $t,m\notin \{j,k\}$.
	By Lemma~\ref{lem:2cvars_either}, it follows that either $t\in \{j,k\}$ (and hence $t = k$) or $m\in\{j,k\}$ (and hence $m = j$).
	Thus, we reach the desired conclusion: there exists $k\in[i]$ such that $k\in S$ for all $\mathcal{S}(\vx_S)\in\mathbf{s}_i$, completing the proof.
\end{proof}

With the help of Lemma~\ref{lem:2cvars_either}, we can prove the following in relation to item (3) on the above list:

\begin{lemma}
	\label{lem:2cvars}
	All CStree stagings of level $i$ in which all stages $\mathcal{S}(\vx_S)$ have $|S| = 2$ are of the form \[ \mathbf{s}_i = \{\mathcal{S}(x_{j_k}x_k) : x_k \in\mathcal{X}_k, x_{j_k}\in\mathcal{X}_{j_k}\} \] for some $j_1,\ldots, j_{d_k}\in[i-1]$ (possibly drawn with repetition) for a given $k\in[i]$.
	The number of stagings of level $i$ of this type is \[ \binom{i}{2} + \sum_{k=1}^i((i-1)^{d_k} - (i - 1)).
	\]
\end{lemma}

\begin{proof}
	By Lemma~\ref{lem:2cvars_commonk}, we know that for a staging $\mathbf{s}_i$ of level $i$ in which all stages $\mathcal{S}(\vx_S)$ have $|S| = 2$ there exists a $k\in[i]$ such that $\vx_{S\cap\{k\}} = x_k$ for some $x_k\in\mathcal{X}_k$ for all stages $\mathcal{S}(\vx_S)\in\mathbf{s}_i$.
	Since all nodes must be contained in a stage in $\mathbf{s}_i$, it follows that for each outcome $x_k\in\mathcal{X}_k$ there is at least one stage $\mathcal{S}(\vx_S)$ in $\mathbf{s}_i$ satisfying $\vx_{S\cap\{k\}} = x_k$.
	Moreover, since every stage has a set of context variables $S$ satisfying $|S| = 2$, we have that $|S\setminus\{k\}| = 1$.
	So we need to consider all possibilities for the additional element of $S$ for each stage.
	We consider them as grouped by their outcome $x_k^\ast$ in their stage-defining context.
	The set of outcomes in $\mathcal{X}_{[i]}$ satisfying $X_k = x_k^\ast$ corresponds to the set of outcomes $\mathcal{X}_{[i]\setminus\{k\}}$.
	Hence, the set of possible CStree stagings in which each stage satisfies $|S| = 2$ and $\vx_{S\cap\{k\}} = x_k^\ast$ corresponds to the possible CStree stagings of $\mathcal{X}_{[i]}$ in which each stage has only one variable in its set of stage-defining contexts.
	By Lemma~\ref{lem:1cvar}, there are exactly $i-1$ such stagings and they correspond to picking an element $j_k\in [i]\setminus k$ and taking the staging of $\mathcal{X}_{[i-1]\setminus k}$ \[ \{\mathcal{S}(x_{j_k}) : x_{j_k} \in\mathcal{X}_{j_k}\}.
	\]
	Thus, any staging of level $i$ in which all stages $\mathcal{S}(\vx_S)$ have $|S| = 2$ is of the form
	\[
		\mathbf{s}_i = \{\mathcal{S}(x_{j_k}x_k) : x_k \in\mathcal{X}_k, x_{j_k}\in\mathcal{X}_{j_k}\}
	\]
	for some $j_1,\ldots, j_{d_k}\in[i-1]$ (possibly drawn with repetition) for a given $k\in[i]$.

	Enumerating these stagings of level $i$, we first pick $k$ in one of $i$ possible ways, then we pick an element from $[i-1]$ for each $\mathcal{X}_k$.
	The number of such choices for a fixed $x$ is equal to the number of functions $\mathcal{X}_k \longrightarrow [i-1]$, which is $(i-1)$.
	Summing over all choices for $k\in[i]$, this yields \[ \sum_{k=1}^i(i-1)^{d_k}.
	\]
	Note however that this sum counts certain stagings twice.
	In the above count we assume the choice of $k$ is the choice of the variable $k$ from Lemma~\ref{lem:2cvars_commonk} that appears in the set of context variables of every stage in the staging.
	However, we include in the above count, the stagings where every stage has exactly the same set of context variables; i.e., $S = \{j,k\}$ for all stages in the staging.
	Since both $j$ and $k$ are considered in the above sum, we count each such staging exactly twice.
	Note that, for fixed $k$, these stagings correspond to the constant functions on $\mathcal{X}_k \longrightarrow [i-1]$, of which there are exactly $(i-1)$ included in each summand in the above sum.
	To avoid overcounting, we thus subtract $(i-1)$ for each summand.
	To count the stagings that satisfy $S = \{j,k\}$ for all stages exactly once, we note that each such staging corresponds to choosing exactly a two element set from $[i]$.
	Hence, the total count of stagings of this type is \[ \binom{i}{2} + \sum_{k=1}^i((i-1)^{d_k} - (i - 1)), \] completing the proof.
\end{proof}

The following lemma enumerates the stagings in item (4) in the above list.

\begin{lemma}
	\label{lem:blendedcvars}
	All CStree stagings of level $i$ that contain at least one stage having context variables $S$ satisfying $|S| = 1$ and at least one stage having context variables $S$ satisfying $|S| = 2$ are of the form \[ \mathbf{s}_i = \{\mathcal{S}(x_k) : x_k\in \mathcal{X}_{k}\setminus K \} \cup \{\mathcal{S}(x_{j_{x_k}}x_k) : x_k\in K, x_{j_{x_k}}\in\mathcal{X}_{j_{x_k}}\} \] for some $k\in [i]$, some nonempty proper subset $K$ of $\mathcal{X}_k$ and some multiset $\{j_{x_k}\in[i]\setminus k : x_k\in K\}$.
	Moreover, the number of stagings of level $i$ of this type is \[ \sum_{k=1}^i(i^{d_k} - (i-1)^{d_k} -1).
	\]
\end{lemma}

\begin{proof}
	Suppose that $\mathbf{s}_i$ is a CStree staging of level $i$ that contains at least one stage having context variables $S$ satisfying $|S| = 1$ and at least one stage having context variables $S$ satisfying $|S| = 2$.
	Let $\mathcal{S}(x_k)$ be the former of the two stages.
	It follows that all other stages $\mathcal{S}(\vx_S)\in\mathbf{s}_i$ satisfy $\vx_{S\cap\{k\}} = x_k^\prime$ for some $x_k^\prime \neq x_k$.
	Otherwise, we would clearly have $\mathcal{S}(\vx_S) \cap\mathcal{S}(x_k)\neq \emptyset$, contradicting the assumption that $\mathbf{s}_i$ is a staging.
	Let $\mathcal{S}(x_jx_k^\prime)$ for some $x_k^\prime \neq x_k$ in $\mathcal{X}_k$ denote the latter stage (e.g. the stage with $|S| = 2$).
	It follows from the proof of Lemma~\ref{lem:2cvars} that \[ \{\mathbf{S}(x_jx_k^\prime) : x_j\in\mathcal{X}_j\}\subset \mathbf{s}_i.
	\]
	Given these restrictions on the stagings of the desired type, we can then produce them all as follows:
	Choose a $k\in[i]$.
	Then choose a nonempty, proper subset $K$ of $\mathcal{X}_k$.
	For each $x_k\in K$, choose $j_{x_k}\in[i]\setminus k$.
	Then \[ \{\mathcal{S}(x_k) : x_k\in \mathcal{X}_{k}\setminus K \} \cup \{\mathcal{S}(x_{j_{x_k}}x_k) : x_k\in K, x_{j_{x_k}}\in\mathcal{X}_{j_{x_k}}\} \] is a CStree staging of the desired type.
	The above argument shows that all stagings of the desired type are of this form.
	Moreover, there are clearly \[ \sum_{k=1}^i\sum_{\emptyset\subsetneq K\subsetneq \mathcal{X}_k}(i-1)^{|K|} \] such stagings of level $i$.
	With the help of the Binomial Theorem, we may rewrite this count as
	\begin{equation*}
		\begin{split}
			\sum_{k=1}^i\sum_{\emptyset\subsetneq K\subsetneq \mathcal{X}_k}(i-1)^{|K|} & = \sum_{k=1}^i\left(\left(\sum_{j=1}^{d_k}\binom{d_k}{j}(i-1)^j\right) - (i-1)^{d_k} - 1\right), \\ &= \sum_{k=1}^i\left((i-1 + 1)^{d_k} - (i-1)^{d_k} - 1\right),\\ &= \sum_{k=1}^i(i^{d_k} - (i-1)^{d_k} -1).
		\end{split}
	\end{equation*}
\end{proof}

\begin{theorem}
	\label{thm:stagingsleveli}
	The CStree stagings of level $i$ in which each stage has at most two context variables are enumerable (i.e., can be listed without redundancy).
	Moreover, there are \[ 1 - \binom{i}{2} + \sum_{k=1}^ii^{d_k} \] many such stagings, where $d_k = |\mathcal{X}_k|$ for $k = 1,\ldots, p$.
\end{theorem}

\begin{proof}
	The enumerability of the stagings follows from combining Lemmas~\ref{lem:0cvars},~\ref{lem:1cvar},~\ref{lem:2cvars} and~\ref{lem:blendedcvars}.
	The total number of such stagings $Z_i$ also follows from this lemma and the following simplification:
	\begin{equation*}
		\begin{split}
			Z_i & = 1 + i + \binom{i}{2} + \sum_{k=1}^i((i-1)^{d_k} - (i - 1)) + \sum_{k=1}^i(i^{d_k} - (i-1)^{d_k} -1), \\ &= 1 + i + \binom{i}{2} - i^2 + \sum_{k=1}^ii^{d_k},\\ &= 1 - \binom{i}{2} + \sum_{k=1}^ii^{d_k}.
		\end{split}
	\end{equation*}
\end{proof}

In the binary setting (i.e., $d_1 = \cdots = d_p)$, Theorem~\ref{thm:stagingsleveli} answers one question recently posed in the combinatorics literature by \citet{alon2023partitioning}.
\citet{alon2023partitioning} are interested in counting the number of ways to partition a $d$-dimensional cube $[0,1]^d$ into disjoint faces.
They note that the general question is difficult and study the subproblem of identify the ways to partition the $d$-cube into disjoint faces where the dimension of each face used in the partition is at most $m$ for a fixed $0\leq m \leq d$.
They denote the number of such partitions by $f_{\leq m}(d)$, and provide some asymptotic estimates for $f_{\leq m}(d)$ for some values of $m$; for instance, in the case when $m = 2$.
In this context, the number of CStree stagings of level $d$ can be viewed as $f_{\geq d - 2}(d)$; that is, the number of partitions of the $d$-cube into disjoint faces of dimension at least $d-2$.
Equivalently, this is the number of partitions of the $d$-dimensional cross-polytope into disjoint faces of dimension at most $2$ (see, for instance, \citep{beck2007computing}).
We summarize this observation in the following corollary.

\begin{corollary}
	\label{cor:cubepartitions}
	The number of partitions of a $d$-dimensional cube into disjoint faces of dimension at least $d-2$ is \[ f_{\geq d-2}(d) = 1 - \binom{i}{2} + i^3.
	\]
\end{corollary}

\begin{proof}
	The result follows from the above discussion and setting $d_1 = \cdots = d_p = 2$ in Theorem~\ref{thm:stagingsleveli}.
\end{proof}

We further note that the Theorem~\ref{thm:stagingsleveli} gives us a formula for the total number of CStrees on $(X_1,\ldots, X_p)$ in which each stage is defined by at most two context variables with a fixed variable ordering, and consequently a formula for the total number of CStrees.
\begin{corollary}
	\label{cor:totalnumtree_fixedorder}
	The number of CStrees on variables $(X_1,\ldots, X_p)$ for fixed $d_1,\ldots, d_p > 1$ and fixed ordering $\pi = 1\cdots p$ having stages that use at most two context variables is \[ \prod_{i = 1}^{p-1}(1 - \binom{i}{2} + \sum_{k=1}^ii^{d_{k}}).
	\]
\end{corollary}

\begin{proof}
	The product formula follows from Theorem~\ref{thm:stagingsleveli}, the observation that specifying a CStree for variables $(X_1,\ldots,X_p)$ requires us to stage $p-1$ levels and the fact that the staging of each level can be done independently from the staging of all others.
\end{proof}

\begin{corollary}
	\label{cor:totalnumtree}
	The number of CStrees on variables $(X_1,\ldots, X_p)$ for fixed $d_1,\ldots, d_p > 1$ having stages that use at most two context variables is \[ \sum_{\pi = \pi_1\cdots\pi_p \in\mathfrak{S}_p}\prod_{i = 1}^{p-1}(1 - \binom{i}{2} + \sum_{k=1}^ii^{d_{\pi_k}}).
	\]
\end{corollary}

\begin{proof}
	This is immediate from the formula in Corollary~\ref{cor:totalnumtree_fixedorder}, which we then sum over all possible orderings of $1,\ldots, p$.
	These are the permutations $\mathfrak{S}_p$.
\end{proof}

It follows from Corollary~\ref{cor:totalnumtree} that an exact search method for learning an optimal CStree would quickly become infeasible for large $p$.
This motivates our choice to learn an optimal ordering, and then perform an exact search over the models for the learned ordering.

For the purposes of analyzing the complexity of the structure learning algorithm presented in \textcolor{cyan}{Section~5}, it is also helpful to know the maximum number of stages possible in a staging of level $i$ of a CStree of the type studied here.
% Section~\ref{sec:learning},
This quantity is given in the following corollary.

\begin{corollary}
	\label{cor:maxpossstages}
	The maximum number of stages in a staging of level $i$ which only has stages defined by contexts using at most two context variables is $d_{(p-1)}d_{(p)}$ where $d_{(k)}$ denotes the $k$-th order statistic on $(d_1,\ldots, d_i)$ and $d_k = |\mathcal{X}_k|$ for all $k\in[i]$.
\end{corollary}

\begin{proof}
	To prove the result we consider the maximum number of stages in a staging of level $i$ as broken down by the list of four items at the start of this section.
	For (1), there is exactly staging and it contains exactly one stage by Lemma~\ref{lem:0cvars}.
	For (2), it follows from Lemma~\ref{lem:1cvar} that each stage of this type contains exactly $d_k$ stages.

	For (3), note that a staging of level $i$ as in Lemma~\ref{lem:2cvars} contains $\sum_{i = 1}^{d_k}d_{j_k}$ stages.
	Hence, if $d_{(p)} = \max(d_1,\ldots, d_p)$ and $d_{(p-1)}$ is the second largest value in $d_1,\ldots, d_p$ (e.g. the $(p-1)$-st order statistic on $d_1,\ldots,d_p$), then the maximum number of stagings of level $i$ of this type is $d_{(1)}d_{(2)}$.

	For (4), we note that each stage having exactly one context variable is refined by two stages having two context variables, and doing this for each stage with one context variable always results in a staging of type (3).
	Hence, the stagings captured in (4) have necessarily fewer stages than the stagings captured in (3).
	We conclude that the maximum number of stages in a staging of level $i$ having only has stages defined by contexts using at most two context variables is $d_{(p-1)}d_{(p)}$.
\end{proof}

\begin{remark}
	\label{rem:appAlpha}
	% \paragraph{Remark 2.}
	The constraint-based phase of the structure learning algorithm \textcolor{cyan}{Algorithm~1} in \textcolor{cyan}{Section~5} reduces the possible context variables that may be used in the stage-defining context $\vx_S$ of the stages $\mathcal{S}(\vx_S)$ in a staging $\mathbf{s}_i$ of level $i$ in the learned CStree.
	% (Algorithm~\ref{alg:cstreelearn})% Section~\ref{sec:learning}
	In principle, this is an optional phase, but it is recommended for efficiency purposes when estimating larger systems.
	In this phase, one also has the option to include the classic DAG sparsity constraint $\alpha$ in \textcolor{cyan}{Assumption~2} commonly used in DAG, LDAG and staged tree learning algorithms.
	As described in the remainder of this remark, such restrictions are easily accommodated into the algorithm by applying \textcolor{cyan}{Theorem~1} to the subsets $K_i$.
	The context-specific sparsity constraint in \textcolor{cyan}{Assumption~1} can then be directly applied without any further considerations.

	Let $K_i\subseteq[i]$ be the set of possible context variables learned for $X_{i+1}$ in the constraint-based phase.
	To apply the above enumeration under this restriction, we simply enumerate the stagings of level $i$ where we restrict the possible choices of context variables from $[i]$ to $K_i$.
	For instance, the number of CStee stagings of level $i$ under this restriction is \[ 1 - \binom{|K_i|}{2} + \sum_{k\in K_i}|K_i|^{d_k}.
	\]
	Bounding the size of the sets of possible context variables $|K_i|\leq \alpha$ for some $\alpha$ therefore significantly decreases the number of CStrees to be considered in the learning process as $p$ grows.
\end{remark}

\begin{remark}
	\label{rem:appBeta}
	% \paragraph{Remark 3.}
	The second sparsity constraint used in our method is bounding the number of context variables used in the stage-defining contexts of each stage.
	The second place where sparsity constraints are introduced in the structure learning algorithm \textcolor{cyan}{Algorithm~1} is when it bounds the cardinality of the sizes of context-variables $|S|$ for each stage $\mathcal{S}(\vx_S)$ in the CStrees.
	% (Algorithm~\ref{alg:cstreelearn})
	An enumeration for all such possible stagings is needed in order to conduct the exact search phase of the algorithm.
	Hence, we necessarily need to solve this combinatorial enumeration problem (or equivalently in the binary case, the problem of computing $f_{\geq d-\beta}(d)$ where we bound $|S| \leq \beta$, as discussed in Corollary~\ref{cor:cubepartitions}).
	Using the enumeration in Theorem~\ref{thm:stagingsleveli}, the current version of \textcolor{cyan}{Algorithm~1} bounds with $\beta=2$.
	% Algorithm~\ref{alg:cstreelearn}
	If we instead restrict to $|S|\leq 1$, we would need only the enumeration given by Lemmas~\ref{lem:0cvars} and~\ref{lem:1cvar}.
	To bound $|S| \leq \beta$ for $\beta >2$, one would need to generalize the above results to an enumeration for the possible stagings of level $i$ for the desired $\beta$.
	We note that, in the binary case, this amounts to solving a version of the problem of \citet{alon2023partitioning} for each $\beta$; i.e., we would necessarily compute $f_{\geq d-\beta}(d)$.
	Notice that increasing either $\beta$ from \textcolor{cyan}{Assumption~1} will allow for learning denser models at the expense of a longer runtime.
\end{remark}

\begin{remark}
	\label{rem:appNotSoSparse}
	% \paragraph{Remark 4.}
	Bounding the size of possible context variables by $\alpha$ (discussed in Remark~\ref{rem:appAlpha} and \textcolor{cyan}{Section~4}) corresponds to a sparsity constraint in the classical DAG setting; that is, it bounds the number of possible parents in a DAG I-MAP of the CStree model.
	The second sparsity bound $\beta$ (as discussed in Remark~\ref{rem:appBeta} and \textcolor{cyan}{Section~4}) only bounds the size of the number of context variables.
	While being a context-specific analogy to bounding the number of parents in a DAG, it is possible to identify CStree models where $\beta = 2$ but the DAG I-MAP of the model is a complete DAG.
	For instance, the CStree model with staged tree representation
	\begin{center}
		\begin{tikzpicture}[thick,scale=0.15] \node[draw, fill=black!
				%---NODES---
				0, inner sep=2pt, rounded corners, minimum width=2pt] (w3) at (6,15)  {\scriptsize 1111};
			\node[draw, fill=black!0, inner sep=2pt, rounded corners, minimum width=2pt] (w4) at (6,13.5) {\scriptsize 1110};
			\node[draw, fill=black!0, inner sep=2pt, rounded corners, minimum width=2pt] (w5) at (6,12) {\scriptsize 1101};
			\node[draw, fill=black!0, inner sep=2pt, rounded corners, minimum width=2pt] (w6) at (6,10.5) {\scriptsize 1100};
			\node[draw, fill=black!0, inner sep=2pt, rounded corners, minimum width=2pt] (v3) at (6,9)  {\scriptsize 1011};
			\node[draw, fill=black!0, inner sep=2pt, rounded corners, minimum width=2pt] (v4) at (6,7.5) {\scriptsize 1010};
			\node[draw, fill=black!0, inner sep=2pt, rounded corners, minimum width=2pt] (v5) at (6,6) {\scriptsize 1001};
			\node[draw, fill=black!0, inner sep=2pt, rounded corners, minimum width=2pt] (v6) at (6,4.5) {\scriptsize 1000};
			\node[draw, fill=black!0, inner sep=2pt, rounded corners, minimum width=2pt] (w3i) at (6,3)  {\scriptsize 0111};
			\node[draw, fill=black!0, inner sep=2pt, rounded corners, minimum width=2pt] (w4i) at (6,1.5) {\scriptsize 0110};
			\node[draw, fill=black!0, inner sep=2pt, rounded corners, minimum width=2pt] (w5i) at (6,0) {\scriptsize 0101};
			\node[draw, fill=black!0, inner sep=2pt, rounded corners, minimum width=2pt] (w6i) at (6,-1.5) {\scriptsize 0100};
			\node[draw, fill=black!0, inner sep=2pt, rounded corners, minimum width=2pt] (v3i) at (6,-3)  {\scriptsize 0011};
			\node[draw, fill=black!0, inner sep=2pt, rounded corners, minimum width=2pt] (v4i) at (6,-4.5) {\scriptsize 0010};
			\node[draw, fill=black!0, inner sep=2pt, rounded corners, minimum width=2pt] (v5i) at (6,-6) {\scriptsize 0001};
			\node[draw, fill=black!0, inner sep=2pt, rounded corners, minimum width=2pt] (v6i) at (6,-7.5) {\scriptsize 0000};

			\node[draw, fill=blue!40, inner sep=2pt, rounded corners, minimum width=2pt] (w1) at (-2,14.25) {\scriptsize 111};
			\node[draw, fill=orange!90, inner sep=2pt, rounded corners, minimum width=2pt] (w2) at (-2,11.25) {\scriptsize 110};
			\node[draw, fill=blue!40, inner sep=2pt, rounded corners, minimum width=2pt] (v1) at (-2,8.25) {\scriptsize 101};
			\node[draw, fill=orange!90, inner sep=2pt, rounded corners, minimum width=2pt] (v2) at (-2,5.25) {\scriptsize 100};
			\node[draw, fill=yellow!90, inner sep=2pt, rounded corners, minimum width=2pt] (w1i) at (-2,2.25) {\scriptsize 011};
			\node[draw, fill=yellow!90, inner sep=2pt, rounded corners, minimum width=2pt] (w2i) at (-2,-0.75) {\scriptsize 010};
			\node[draw, fill=red!90, inner sep=2pt, rounded corners, minimum width=2pt] (v1i) at (-2,-3.75) {\scriptsize 001};
			\node[draw, fill=red!90, inner sep=2pt, rounded corners, minimum width=2pt] (v2i) at (-2,-6.75) {\scriptsize 000};

			\node[draw, fill=green!0, inner sep=2pt, rounded corners, minimum width=2pt] (w) at (-8,12.75) {\scriptsize 11};
			\node[draw, fill=cyan!0, inner sep=2pt, rounded corners, minimum width=2pt] (v) at (-8,6.75) {\scriptsize 10};
			\node[draw, fill=green!0, inner sep=2pt, rounded corners, minimum width=2pt] (wi) at (-8,0.75) {\scriptsize 01};
			\node[draw, fill=cyan!0, inner sep=2pt, rounded corners, minimum width=2pt] (vi) at (-8,-5.25) {\scriptsize 00};

			\node[draw, fill=yellow!0, inner sep=2pt, rounded corners, minimum width=2pt] (r) at (-14,9.75) {\scriptsize 1};
			\node[draw, fill=yellow!0, inner sep=2pt, rounded corners, minimum width=2pt] (ri) at (-14,-1.75) {\scriptsize 0};

			\node[draw, fill=black!0, inner sep=2pt, rounded corners, minimum width=2pt] (I) at (-20,3) {\scriptsize r};

			%---EDGES---
			\draw[->]   (I) --    (r) ;
			\draw[->]   (I) --   (ri) ;

			\draw[->]   (r) --   (w) ;
			\draw[->]   (r) --   (v) ;

			\draw[->]   (w) --  (w1) ;
			\draw[->]   (w) --  (w2) ;

			\draw[->]   (w1) --   (w3) ;
			\draw[->]   (w1) --   (w4) ;
			\draw[->]   (w2) --  (w5) ;
			\draw[->]   (w2) --  (w6) ;

			\draw[->]   (v) --  (v1) ;
			\draw[->]   (v) --  (v2) ;

			\draw[->]   (v1) --  (v3) ;
			\draw[->]   (v1) --  (v4) ;
			\draw[->]   (v2) --  (v5) ;
			\draw[->]   (v2) --  (v6) ;

			\draw[->]   (ri) --   (wi) ;
			\draw[->]   (ri) -- (vi) ;

			\draw[->]   (wi) --  (w1i) ;
			\draw[->]   (wi) --  (w2i) ;

			\draw[->]   (w1i) --  (w3i) ;
			\draw[->]   (w1i) -- (w4i) ;
			\draw[->]   (w2i) --  (w5i) ;
			\draw[->]   (w2i) --  (w6i) ;

			\draw[->]   (vi) --  (v1i) ;
			\draw[->]   (vi) --  (v2i) ;

			\draw[->]   (v1i) --  (v3i) ;
			\draw[->]   (v1i) -- (v4i) ;
			\draw[->]   (v2i) -- (v5i) ;
			\draw[->]   (v2i) --  (v6i) ;

			%---LABELS---
			\node at (-17.5,-9) {$X_1$} ;
			\node at (-11.5,-9) {$X_2$} ;
			\node at (-5,-9) {$X_3$} ;
			\node at (2,-9) {$X_4$} ;

		\end{tikzpicture}
	\end{center}
	has the (minimal) DAG I-MAP
	\begin{center}
		\begin{tikzpicture}[thick,scale=0.2]

			\node[circle, draw, fill=black!
				0, inner sep=1pt, minimum width=1pt] (H1) at (3.25,8) {$1$};
			\node[circle, draw, fill=black!0, inner sep=1pt, minimum width=1pt] (B1) at (-2.25,4) {$2$};
			\node[circle, draw, fill=black!0, inner sep=1pt, minimum width=1pt] (G1) at (8.25,4) {$3$};
			\node[circle, draw, fill=black!0, inner sep=1pt, minimum width=1pt] (B2) at (3.25,0) {$4$};

			%---EDGES---
			\draw[->]   (H1) -- (B1) ;
			\draw[->]   (H1) -- (G1) ;
			\draw[->]   (H1) -- (B2) ;
			\draw[->]   (B1) -- (B2) ;
			\draw[->]   (G1) -- (B2) ;
			\draw[->]   (B1) -- (G1) ;
		\end{tikzpicture}
	\end{center}
	This can be verified using the method described in Section~\ref{subsec:LDAGrep}.
	The choice of value for the bound $\beta$ does inevitably impact sparsity of the DAG but this depends on the cardinalities of the state spaces of the variables being models.
	For instance, when $\beta = 2$, and all variables are binary, it follows from Lemma~\ref{lem:2cvars} that the maximum number of parents in the DAG I-MAP of the CStree is $3$.
	This corresponds to the maximum number of stages (i.e., $4$) from Corollary~\ref{cor:maxpossstages}, but distrributed in a non-uniform way as in the example here.
	More generally, if the all variables have state space of cardinality $d$ then the maximum number of parents is $d+1$.
\end{remark}

\section{Additional experimental results}
\label{sec:additionalexp}

\subsection{Scalability Analysis}
We include a second plot with the same experimental set up as the plot in \textcolor{cyan}{Figure~3b}, but for $n = 10000$ samples.
% Figure~\ref{fig:runtime1000},
In particular, we note that the sample size does not have a significant effect on the runtime.
In our implementation of \textcolor{cyan}{Algorithm~1}, the data is only used when computing the context marginal likelihoods \textcolor{cyan}{(7)}.
% Algorithm~\ref{alg:cstreelearn},% \eqref{eqn:contextmarginallikelihood},
These are computed with the help of \texttt{pandas}, which can efficiently perform the necessary computations.
\begin{center}
	\includegraphics[width=0.45\textwidth]{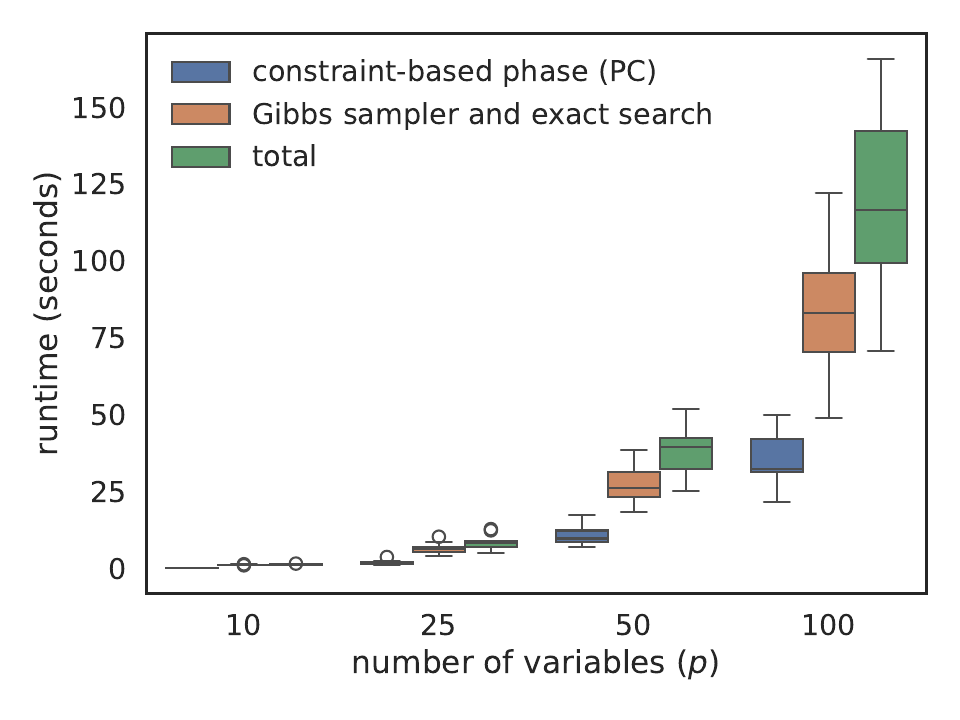}
\end{center}

All reported scalability results were run on a c5a.8xlarge instance from Amazon EC2.

\subsection{Real World Example: ALARM data set}
The ALARM data set analyzed in \textcolor{cyan}{Subsection~6.3} is available as part of the \texttt{bnlearn} package in \texttt{R}.
% Subsection~\ref{subsec:realdata}
The variables and the state spaces for the data set are listed in Table~\ref{table:alarmvars}.
We identify the states of each variable with $0,1,2,3$ from left-to-right as they are listed in the column of Table~\ref{table:alarmvars}.
Using Table~\ref{table:alarmvars}, we can interpret the 12 labels the edges in \textcolor{cyan}{Figure~4}.
% Figure~\ref{fig:realdata}.
The corresponding CSI relations for each label are listed in Table~\ref{table:alarmCSIs}.
The labels in Table~\ref{table:alarmCSIs} offer some insights into what the data shows at a context-specific level that cannot be encoded in a DAG model.
For instance the label $\mathcal{L}(32\rightarrow 13) = \{2,3\}$ indicates that arterial $CO_2$ levels are independent of expelled $CO_2$ levels given that lung ventilation is at least a normal level.

\begin{table}[ht]
	\begin{tabular}{|l | l | l |}\hline
		{\bf Label}            & {\bf CSI relations}                          & {\bf CSI relations expressed in real variables}                                 \\\hline
		$L_{29, 7} = \{2\}$    & $X_7\independent X_{29} \mid X_{28} = 2$     & $\textrm{HREK} \independent \textrm{ERCA} \mid \textrm{HR} = \textrm{high}$     \\\hline

		$L_{29, 8} = \{2\}$    & $X_{8} \independent X_{29} \mid X_{28} = 2$  & $\textrm{HRSA} \independent \textrm{ERCA} \mid \textrm{HR} = \textrm{high}$     \\\hline

		$L_{30, 10} = \{1\}$   & $X_{10} \independent X_{30} \mid X_{31} = 1$ & $\textrm{SAO2} \independent \textrm{SHNT} \mid \textrm{PVS} = \textrm{normal}$  \\\hline

		$L_{32, 13} = \{2,3\}$ & $X_{13} \independent X_{32} \mid X_{34} = 2$ & $\textrm{ECO2} \independent \textrm{ACO2} \mid \textrm{VLNG} = \textrm{normal}$ \\\hline
		                       & $X_{13} \independent X_{32} \mid X_{34} = 3$ & $\textrm{ECO2} \independent \textrm{ACO2} \mid \textrm{VLNG} = \textrm{high}$   \\\hline

		$L_{21, 14} = \{3\}$   & $X_{14} \independent X_{21} \mid X_{34} = 3$ & $\textrm{MINV} \independent \textrm{INT} \mid \textrm{VLNG} = \textrm{high}$    \\\hline

		$L_{16, 24} = \{1\}$   & $X_{24} \independent X_{16} \mid X_{17} = 1$ & $\textrm{LVV} \independent \textrm{HYP} \mid \textrm{LVF} = \textrm{true}$      \\\hline

		$L_{16, 25} = \{1\}$   & $X_{25} \independent X_{16} \mid X_{17} = 1$ & $\textrm{STKV} \independent \textrm{HYP} \mid \textrm{LVF} = \textrm{true}$     \\\hline

		$L_{10, 26} = \{1\}$   & $X_{26} \independent X_{10} \mid X_{3} = 1$  & $\textrm{CCHL} \independent \textrm{SAO2} \mid \textrm{TPR} = \textrm{normal}$  \\\hline

		$L_{21, 30} = \{1\}$   & $X_{30} \independent X_{21} \mid X_{20} = 1$ & $\textrm{SHNT} \independent \textrm{INT} \mid \textrm{PMP} = \textrm{true}$     \\\hline

		$L_{26, 32} = \{0,3\}$ & $X_{32} \independent X_{26} \mid X_{33} = 0$ & $\textrm{ACO2} \independent \textrm{CCHL} \mid \textrm{VALV} = \textrm{zero}$   \\\hline
		                       & $X_{32} \independent X_{26} \mid X_{33} = 3$ & $\textrm{ACO2} \independent \textrm{CCHL} \mid \textrm{VALV} = \textrm{high}$   \\\hline

		$L_{11, 31} = \{2,3\}$ & $X_{31} \independent X_{11} \mid X_{33} = 2$ & $\textrm{PVS} \independent \textrm{FIO2} \mid \textrm{VALV} = \textrm{normal}$  \\\hline
		                       & $X_{31} \independent X_{11} \mid X_{33} = 3$ & $\textrm{PVS} \independent \textrm{FIO2} \mid \textrm{VALV} = \textrm{high}$    \\\hline
		$L_{23, 35} = \{2,3\}$ & $X_{35} \independent X_{23} \mid X_{36} = 2$ & $\textrm{VTUB} \independent \textrm{DISC} \mid \textrm{VMCH} = \textrm{normal}$ \\\hline
		                       & $X_{35} \independent X_{23} \mid X_{36} = 3$ & $\textrm{VTUB} \independent \textrm{DISC} \mid  \textrm{VMCH} = \textrm{high}$  \\\hline
	\end{tabular}
	\caption{CSI relations learned for ALARM data set.}
	\label{table:alarmCSIs}
\end{table}

% maporder: [29, 21, 11, 27, 36, 16, 23, 35, 12, 19, 3, 34, 17, 24, 25, 33, 20, 31, 30, 0, 10, 26, 28, 7, 22, 32, 5, 18, 14, 8, 6, 13, 4, 15, 1, 9, 2]

\begin{table}[ht]
	\begin{tabular}{|l | l | l|}\hline
		{\bf Variable index} & {\bf Variable in ALARM data set}             & {\bf State Space}            \\\hline
		0                    & CVP (central venous pressure)                & Low, Normal, High            \\\hline
		1                    & PCWP (pulmonary capillary wedge pressure)    & Low, Normal, High            \\\hline
		2                    & HIST (history)                               & False, True                  \\\hline
		3                    & TPR (total peripheral resistance)            & Low, Normal, High            \\\hline
		4                    & BP (blood pressure)                          & Low, Normal, High            \\\hline
		5                    & CO (cardiac output)                          & Low, Normal, High            \\\hline
		6                    & HRBP (heart rate / blood pressure)           & Low, Normal, High            \\\hline
		7                    & HREK (heart rate measured by an EKG monitor) & Low, Normal, High            \\\hline
		8                    & HRSA (heart rate / oxygen saturation)        & Low, Normal, High            \\\hline
		9                    & PAP (pulmonary artery pressure)              & Low, Normal, High            \\\hline
		10                   & SAO2 (arterial oxygen saturation)            & Low, Normal, High            \\\hline
		11                   & FIO2 (fraction of inspired oxygen)           & Low, Normal, High            \\\hline
		12                   & PRSS (breathing pressure)                    & Zero, Low, Normal, High      \\\hline
		13                   & ECO2 (expelled CO2)                          & Zero, Low, Normal, High      \\\hline
		14                   & MINV (minimum volume)                        & Zero, Low, Normal, High      \\\hline
		15                   & MVS (minimum volume set)                     & Low, Normal, High            \\\hline
		16                   & HYP (hypovolemia)                            & False, True                  \\\hline
		17                   & LVF (left ventricular failure)               & False, True                  \\\hline
		18                   & APL (anaphylaxis)                            & False, True                  \\\hline
		19                   & ANES (insufficient anesthesia/analgesia)     & False, True                  \\\hline
		20                   & PMB (pulmonary embolus)                      & False, True                  \\\hline
		21                   & INT (intubation)                             & Normal, Esophageal, Onesided \\\hline
		22                   & KINK (kinked tube)                           & False, True                  \\\hline
		23                   & DISC (disconnection)                         & False, True                  \\\hline
		24                   & LVV (left ventricular end-diastolic volume)  & Low, Normal, High            \\\hline
		25                   & STKV (stroke volume)                         & Low, Normal, High            \\\hline
		26                   & CCHL (catecholamine)                         & Normal, High                 \\\hline
		27                   & ERLO (error low output)                      & False, True                  \\\hline
		28                   & HR (heart rate)                              & Low, Normal, High            \\\hline
		29                   & ERCA (electrocauter)                         & False, True                  \\\hline
		30                   & SHNT (shunt)                                 & Normal, High                 \\\hline
		31                   & PVS (pulmonary venous oxygen saturation)     & Low, Normal, High            \\\hline
		32                   & ACO2 (arterial CO2)                          & Low, Normal, High            \\\hline
		33                   & VALV (pulmonary alveoli ventilation)         & Zero, Low, Normal, High      \\\hline
		34                   & VLNG (lung ventilation)                      & Zero, Low, Normal, High      \\\hline
		35                   & VTUB (ventilation tube)                      & Zero, Low, Normal, High      \\\hline
		36                   & VMCH (ventilation machine)                   & Zero, Low, Normal, High      \\\hline
	\end{tabular}
	\caption{Variable for ALARM data set.}
	\label{table:alarmvars}
\end{table}

\clearpage

\subsection{Real Data Example: Sachs Protein Expression Data Set}

While the ALARM data set is meant to model a real world scenario, the data is in fact synthetic.
To give a proper real data example, we run \textcolor{cyan}{Algorithm~1} on the well-studied Sachs protein expression data set \citep{sachs2005causal}.
% Algorithm~\ref{alg:cstreelearn}
The Sachs data set is a standard benchmark data set consisting of 7466 measurements of the abundance of phospholipids and phosphoproteins in primary human immune system cells.
The measurements were taken under various experimental conditions, and the data set is purely interventional in its raw form.
An observational data set can be extracted from the raw data as described in \cite{wang2017permutation}.
The resulting observational version of the data set has 1755 samples from the joint distribution of 11 phosphoproteins.
The samples are purely numerical, but highly non-normal according to Shapiro-Wilks tests performed on the marginal data of each protein.
Hence, it is reasonable to discretize the data and develop a discrete model for the data.
In this case, we binarize each variable by binning according to the upper and lower $50\%$-quantiles.
To demonstrate how \textcolor{cyan}{Algorithm~1} performs when we do not bound the set of possible contexts variables according to an estimated CPDAG, we ran it on this data with the sets $K_i$ set to all variables excluding the variable $i$.
% Algorithm~\ref{alg:cstreelearn}
This is equivalent to setting the bound $\alpha$ from Remark~2 equal to $10$, which is equivalent to $\alpha = \infty$ for a system with only $11$ variables.
When we remove the sparsity constraint $\alpha$, \textcolor{cyan}{Algorithm~1} ran on these $11$ binary variables in approximately $2.6$ minutes.
While this is still a reasonable compute time to search over the $114,561,216,000$ possible CStrees (see Corollary~\ref{cor:totalnumtree}), this shows that the classic DAG sparsity constraint still assists in speedy computation.
The LDAG representation of the estimated CStree is depicted below.
\begin{center}
	\includegraphics[width=0.45\textwidth]{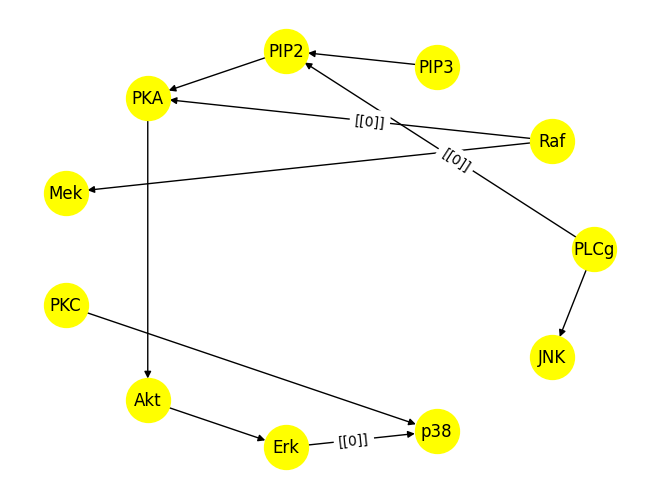}
\end{center}
We see that \textcolor{cyan}{Algorithm~1} learned three CSI relations not captured by a DAG representation of the data.
% Algorithm~\ref{alg:cstreelearn}
The label $L_{\textrm{\scriptsize PCLg}, \textrm{\scriptsize PIP2}} = \{0\}$ encodes $\textrm{PIP2}\independent \textrm{PCLg} \mid \textrm{PIP3} = 0$, meaning that PIP3 and PCLg are independent when the expression level of PIP3 is low.
The label $L_{\textrm{\scriptsize Raf}, \textrm{\scriptsize PKA}} = \{0\}$ encodes $\textrm{PKA} \independent \textrm{Raf} \mid \textrm{PIP2} = 0$, meaning the expression levels of Raf and PKA are independent when the expression of PIP2 is low.
Similarly, the label $L_{\textrm{\scriptsize ERK}, \textrm{\scriptsize p38}} = \{0\}$ indicates that the expression levels of ERK and p38 are independent when the expression level of PKC is low.

\newpage

\subsection{Real Data Example: Mushroom}
\label{supp:mushroom}

The following table gives the contexts specified by the edge labels in Figure~\ref{fig:mushroom-ldag}.
\begin{table}[ht]
	\centering
	\begin{tabular}{| r | l |}
		\hline
		edge label & context                                                                         \\\hline
		a          & [[1]]                                                                           \\\hline
		b          & [[3]]                                                                           \\\hline
		c          & [[0, 1], [1, 1], [3]]                                                           \\\hline
		d          & [[0, 1], [1, 1], [3]]                                                           \\\hline
		e          & [[1], [2]]                                                                      \\\hline
		f          & [[0, 0], [1, 0], [2, 0], [3, 0]]                                                \\\hline
		g          & [[0, 0], [1, 0], [2, 0], [3, 0]]                                                \\\hline
		h          & [[1], [2], [3, 0], [3, 1]]                                                      \\\hline
		i          & [[0, 0], [0, 1], [0, 2], [0, 3]]                                                \\\hline
		j          & [[0, 0], [1, 0], [0, 2], [1, 2], [3]]                                           \\\hline
		k          & [[0, 0], [1, 0], [0, 2], [1, 2], [3]]                                           \\\hline
		l          & [[0, 0], [1, 0], [2, 0], [3, 0]]                                                \\\hline
		m          & [[0, 0], [1, 0], [2, 0], [3, 0]]                                                \\\hline
		n          & [[0, 0], [1, 0], [2, 0], [3, 0]]                                                \\\hline
		o          & [[0, 1], [1, 1], [2, 1], [3, 1], [2], [3], [4], [0, 5], [1, 5], [2, 5], [3, 5]] \\\hline
		p          & [[0, 0], [1, 0], [2, 0], [3, 0], [1], [0, 3], [1, 3], [2, 3], [3, 3], [4], [6]] \\\hline
		q          & [[1, 0], [1, 1]]                                                                \\\hline
		r          & [[0, 1], [1, 1]]                                                                \\\hline
		s          & [[0, 1], [1, 1]]                                                                \\\hline
		t          & [[0, 1], [1, 1]]                                                                \\\hline
		u          & [[0, 1], [1, 1]]                                                                \\\hline
		v          & [[0, 1], [1, 1]]                                                                \\\hline
		w          & [[0, 0], [1, 0], [2], [3], [4]]                                                 \\\hline
		x          & [[1, 0], [1, 1]]                                                                \\\hline
		y          & [[1], [0, 2], [1, 2], [4], [0, 5], [1, 5], [6]]                                 \\\hline
		z          & [[0], [2], [3]]                                                                 \\\hline
		A          & [[1, 0], [1, 1]]                                                                \\\hline
		B          & [[0], [1], [2]]                                                                 \\\hline
		C          & [[0, 0], [0, 1], [0, 2], [0, 3]]                                                \\\hline
		D          & [[0, 0], [1, 0], [1], [2]]                                                      \\\hline
		E          & [[0], [1], [2], [3], [4], [6]]                                                  \\\hline
		F          & [[0, 0], [0, 1]]                                                                \\\hline
	\end{tabular}
	\caption{Edge labels and corresponding contexts for Mushroom LDAG.}
\end{table}

\section{Score table computations}
\label{sec:complexity}

\begin{lemma}
	\label{lem:stagelik}
	The time and space complexity for computing and storing the context marginal likelihoods is $\mathcal O(p {\binom{|K|}{m}} d)$ and $\mathcal O(p {\binom{|K|}{m}} d^m)$, respectively.
\end{lemma}
\begin{proof}

	We denote the set of possible contexts for $X_i$ with variables in $K_i$ by $\mathcal{C}_{K_i} = \bigcup_{S\subset K_i }\{\vx_{S} | \vx_{S}\in \mathcal{X}_S\}$.
	For each $X_i$ and associated contexts $\vx_S \in \mathcal{C}_{K_i}$, we let $s = \mathcal S(\vx_{S})$ be an arbitrary stage defined by the context $\vx_S$ and calculate the context marginal likelihoods $z_{i,{\vx_S} }$ of \textcolor{cyan}{(7)}.
	% \eqref{eqn:contextmarginallikelihood}.
	Assuming the counts $N_{isk}$ in \textcolor{cyan}{(7)} are pre-calculated and can be accessed in $\mathcal{O}(1)$, the context marginal likelihoods can be computed in $\mathcal O(p {\binom{|K|}{m}} d)$ time.
	% \eqref{eqn:contextmarginallikelihood}
	By using a bijective map from each $K_i$ to the integers, the context marginal likelihoods can be accessed with a time complexity $\mathcal O(1)$ and stored with a space complexity of $\mathcal O(p {\binom{|K|}{m}} d^m)$.

\end{proof}

\begin{theorem}
	\label{thm:loscores}
	The time and space complexity for computing and storing the local ordering scores is $\mathcal O(p2^{|K|}|\mathcal S_{K,m}|d^m)$ and $\mathcal O(p2^{|K|})$, respectively.
\end{theorem}
\begin{proof}

	For each $X_i$, and $L \subset K_i$ we compute local ordering scores in \textcolor{cyan}{(8)} using the look-up table for context marginal likelihoods $z_{i,{\vx_S} }$.
	% \eqref{eqn:loscores}
	By Lemma~\ref{lem:stagelik}, the $z_{i,{\vx_S} }$ values are accessed in $\mathcal O(1)$, so the local order scores can be pre-computed in $\mathcal O(p2^{|K|}|\mathcal S_{K,m}|d^m)$, where $2^{|K|}$ is the number of subsets of $K$.
	To see this, note that summing over all stagings contributes with the factor, $|\mathcal S_{K,m}|$, which is enumerated using Corollary~\ref{cor:totalnumtree_fixedorder}.
	From Corollary~\ref{cor:maxpossstages} we have that maximum the number of stages in a staging is $\mathcal O(d^m)$.
	Together with Lemma~\ref{lem:stagelik}, we find a total time complexity of $\mathcal O(p {\binom{|K|}{m}} d + p2^{|K|}|\mathcal S_{K,m}|d^m) = \mathcal O(p2^{|K|}|\mathcal S_{K,m}|d^m) $.
	Using a bijective mapping from the subsets of $K_i$ to the integers, the local ordering scores can be stored with a time complexity of $\mathcal O(p2^{|K|})$ and accessed in $\mathcal{O}(1)$.
\end{proof}

\end{document}